\documentclass[sigconf]{acmart}
\usepackage{graphicx}
\usepackage{bm}
\usepackage{caption, subcaption}
\usepackage{etoolbox}
\usepackage{enumitem}
\usepackage{multicol}
\usepackage{rotating}
\usepackage{hyperref}
\usepackage{cleveref}
\usepackage{float}
\usepackage{multirow}
\usepackage{here}

\AtBeginDocument{%
  \providecommand\BibTeX{{%
    \normalfont B\kern-0.5em{\scshape i\kern-0.25em b}\kern-0.8em\TeX}}}

\setcopyright{none}
\renewcommand\footnotetextcopyrightpermission[1]{}

\makeatletter
\renewcommand\@formatdoi[1]{\ignorespaces}
\makeatother

\begin{document}

\title[Classification Benchmarks for Under-resourced Bengali Language based on Multichannel Convolutional-LSTM Network]{Classification Benchmarks for Under-resourced\\ Bengali Language based on Multichannel\\ Convolutional-LSTM Network}

\acmConference{\vphantom{BLA}  \begingroup\color{white} }{\vphantom{BLA}}{\vphantom{BLA} \endgroup}

\author{Md. Rezaul Karim}
\affiliation{
    \institution{Fraunhofer FIT, Aachen, Germany}
    \institution{RWTH Aachen University, Germany}
}
\author{Bharathi Raja Chakravarthi}
\affiliation{
    \institution{Insight SFI Research Centre for Data Analytics,\\ Data Science Institute, National University of Ireland Galway, Ireland}
}
\author{John P. McCrae}
\affiliation{
    \institution{Insight SFI Research Centre for Data Analytics,\\ Data Science Institute, National University of Ireland Galway, Ireland}
}
\author{Michael Cochez}
\affiliation{
    \institution{Department of Computer Science,\\ Vrije Universiteit Amsterdam, Netherlands}
}
\renewcommand{\shortauthors}{Karim, Chakravarthi, McCrae, and Cochez}

\begin{abstract}
    Exponential growths of social media and micro-blogging sites not only provide platforms for empowering freedom of expressions and individual voices but also enables people to express anti-social behaviour like online harassment, cyberbullying, and hate speech. Numerous works have been proposed to utilize these data for social and anti-social behaviours analysis, document characterization, and sentiment analysis by predicting the contexts mostly for highly resourced languages such as English. However, there are languages that are under-resources, e.g., South Asian languages like Bengali, Tamil, Assamese, Telugu that lack of computational resources for the natural language processing~(NLP)\footnote{List of abbreviations can be found at the end of this paper.} tasks. In this paper\footnote{This paper is under review in the Journal of Natural Language Engineering.}, we provide several classification benchmarks for Bengali, an \texttt{under-resourced language}. We prepared three datasets of expressing hate, commonly used topics, and opinions for hate speech detection, document classification, and sentiment analysis, respectively. We built the largest Bengali word embedding models to date based on 250 million articles, which we call \texttt{BengFastText}. We perform three different experiments, covering document classification, sentiment analysis, and hate speech detection. We incorporate word embeddings into a Multichannel Convolutional-LSTM~(\texttt{MConv-LSTM}) network for predicting different types of hate speech, document classification, and sentiment analysis. Experiments demonstrate that \texttt{BengFastText} can capture the semantics of words from respective contexts correctly. Evaluations against several baseline embedding models, e.g., Word2Vec and GloVe yield up to 92.30\%, 82.25\%, and 90.45\% F1-scores in case of document classification, sentiment analysis, and hate speech detection, respectively during 5-fold cross-validation tests.
\end{abstract}

\keywords{Under-resource language, NLP, Deep learning, Sentiment analysis, Hate speech detection, Word embedding, Word2Vec, FastText.}

\maketitle
\section{Introduction}
In recent years, micro-blogging platforms and social networking sites have grown exponentially, enabling their users to voice their opinions~\cite{khan2018two}. At the same time, they have also enabled anti-social behavior~\cite{hate1}, online harassment, cyberbullying, false political and religious rumor, and hate speech activities~\cite{DBLP:conf/icwsm/RibeiroCSAM18,kshirsagar2018predictive}. The anonymity and mobility offered by such media have made the breeding and spread of hate speech~\cite{zhang2018detecting} -- eventually leading to hate crime in many aspects including religious, political, geopolitical, personal, and gender abuse.
This applies to every human being regardless of languages, geographic locations or ethnicity. While the Web began as an overwhelmingly English phenomenon, it now contains extant text in hundreds of languages. In the real world, languages are becoming extinct at an alarming rate and the world could lose more than half of its linguistic diversity by the year 2100. Certain languages are low resource because they are actually low-density languages based on the number of people speaking in the world. 

Languages are dynamic being spoken by communities whose lives are shaped by the rapidly changing world. 
The number of living languages spoken in the world is about 7,099, which is constantly in flux. About a third of these languages are now endangered, often with less than 1,000 speakers remaining. Meanwhile, only 23 languages account for more than half the world's population. Bengali, which is a rich language with a lot of diversity is spoken in Bangladesh, the second most commonly spoken language in  India, and the seventh most commonly spoken language in the world with nearly 230 million speakers~(200 million native speakers)~\cite{islam2009research}.

Sentiment analysis makes use of the polarity expressed in texts, where lexical resources are often used to look up specific words as their presence can be predictive features. 
Linguistic features, on the other hand, utilize syntactic information like Part of Speech~(PoS) and certain dependency relations as features. 
Content-based classification, which is the process of grouping documents into different classes or categories, is emerging due to the continuous growth of digital data. 
Apart from these, sentiment analysis in Bengali is progressively being considered as a non-trivial task, for which previous approaches have attempted to detect the overall polarity of Bengali texts. 
\texttt{Hate speech}, which is defined by the Encyclopedia of American Constitution \cite{nockleby2000hate} is ``any communication that disparages a person or a group on the basis of some characteristic such as race, color, ethnicity, gender, sexual orientation, nationality, religion, or other characteristics'', is also rampant online, e.g., social media, newspapers, and books~\cite{hate3}. 
Like in other major languages like in English, the use of hate speech in Bengali is pervasive and can have serious consequences, because according to a Special Rapporteur to the UN Humans Rights Council, failure to monitor and react to such hate speech in a timely manner can reinforce the subordination of targeted minorities, making them not only vulnerable to attack but also influencing majority populations and making them more indifferent to various manifestations of such hatred \cite{hate5}.

\begin{figure*}
	\centering
	\includegraphics[width=0.95\textwidth]{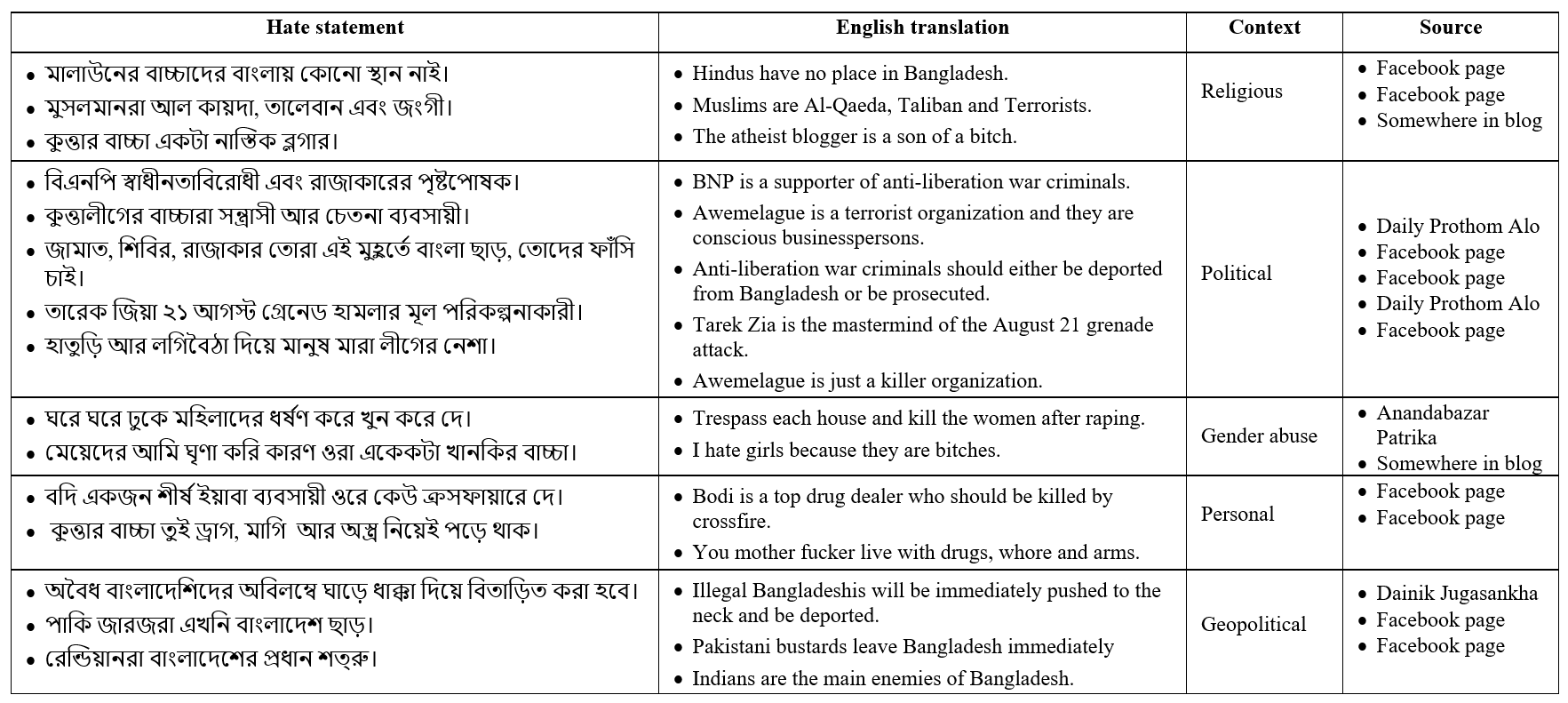}	
    \caption{Examples of Bengali hate speech, either directed towards a specific person or entity, or generalized towards a group}	
	\label{cdec_wf1}
\end{figure*}
\raggedbottom

Thus, identifying such hate speech and making awareness in a virtual landscape beyond the realms of traditional law enforcement is a non-trivial research contribution. 
However, a vast amount of content published online does not allow for manual analysis. For years, tech giants like Twitter, Facebook, and YouTube have been combating the issue of hate speech and it has been estimated that hundreds of millions of euros are being invested every year on counter-measures including manpower~\cite{hate5}.
However, they are still being criticized for not doing enough~\cite{hate4}. 
Efficient and automated techniques for such useful researches for under-resourced Bengali language is in scarce, which is largely because such measures require significant manual efforts. However, reviewing online contents to identify offensive materials and taking actions need manual intervention, which is not only labor intensive and time consuming but also not scalable~\cite{hate5}. 

To the end, computational linguistics has long been dominated by techniques specifically geared towards English and other European languages in which English language models are often used as baselines, despite the fact that it is an outlier in a lot of linguistic aspects. Despite being often unreported if the results hold up in other languages, the techniques are often considered state-of-the-art for natural language understanding. The field now, has unprecedented access to a great number of low-resource languages, readily available to be studied, but needs to act quickly before political, social, and economic pressures cause these languages to disappear from the real and/or virtual worlds. 
In the absence of these resources, the lack of good performance across various tasks feeds into a systemic information asymmetry in the digital world. The quality of translation systems, the lack of social media corpora, and the general absence of language-agnostic tools is going to exacerbate the digital divide in coming times at the global level and is a prominent barrier in the way of truly democratizing technology access in the world. 

Accurate identification of such hate speeches or sentiment analysis requires not only an automated, scalable, and accurate methods but also computational resources such as state-of-the-art machine learning~(ML) approaches, word embedding models and rich labelled datasets, which in turn needs natural language processing~(NLP) resources for the language in question is the core contribution of this work. This is obvious, because the NLP resources for Bengali are still scarce~\cite{BESACIER201485,chakravarthi2018improving}.
Recent research efforts from both the NLP and ML communities have proven to be very effective for well-resourced languages like English. Computational linguistics today is largely dominated by supervised machine learning~(ML), which requires annotating data for low-resource languages that can be time consuming, expensive and needs experts. 
They identified these research problems as supervised learning tasks, which can be divided into two categories.
First, there are classic methods, which rely on manual feature engineering such as Support Vector Machines~(SVM), Na{\"i}ve Bayes~(NB), KNN, Logistic Regression~(LR), Decision Trees~(DT), Random Forest~(RF) and Gradient Boosted Trees(GBT). 
Second, there are approaches which employs deep neural networks~(DNN) to learn multilayers of abstract features from raw texts. These approaches are largely based on Convolutional~(CNN) or Long Short-term Memory~(LSTM) networks. 

These approaches are mostly deep learning~(DL) based, referring to the depth of the neural networks used. Nevertheless, despite the growth of studies and existing ML approaches, under-resourced languages such as Bengali lacks resources such as rich word embedding models and comparative evaluation on NLP tasks. In this work, we built a Bengali word embedding model called \texttt{BengFastText} and implemented a robust neural network architecture called multichannel convolutional-LSTM network~(\texttt{MConv-LSTM}) by combining both CNN and LSTM networks with multiple channels in a single architecture to leverage their benefits. The other key contributions of this paper can be summarized as follows: 

\vspace{1mm}
\begin{itemize}
    \item We built the largest Bengali word embedding model called \texttt{BengFastText} based on 250 million Bengali articles in an under-resourced setting. \texttt{BengFastText} is a language model based on \texttt{fastText}~\cite{fastText}, which is computationally-efficient predictive model for learning word embeddings from raw Bengali texts.

    \item We prepared three public datasets of hate speech, sentiment analysis, and document classification of Bengali texts, which are larger than any currently available datasets by both quantity and subject coverage.

    \item We trained several ML baseline models like SVM, KNN, LR, NB, DT, RF and GBT. Further, we prepared embeddings based on Word2Vec and GloVe for a comparative analysis between ML and \texttt{MConv-LSTM} model. 

    \item Our approach shows how the automatic feature selection using \texttt{MConv-LSTM} model significantly improves the overhead of manual feature engineering.
\end{itemize}
\vspace{1mm}

The rest of the paper is structured as follows: \Cref{rw} reviews related work on hate speech and Bengali word embedding. \Cref{section:3} describes the data collection and annotation process. \Cref{nettwork} describes the process of Bengali neural embedding, network construction, and training. \Cref{er} illustrates experiment results including a comparative analysis with baseline models on all datasets. \Cref{con} summarizes this research with potential limitations and points some possible outlook before concluding the paper.

\section{Related Work}
\label{rw}
Although linguistic analyses are carried out extensively and well studied on other major languages, only a few approaches have been explored for Bengali~\cite{islam2009research}, which is due to lack of systematic method of text collection, annotated corpora, name dictionaries, morphological analyzer, and overall research outlook.
Existing work on Bengali NLP mostly focus on either document categorization~\cite{islam2017comparative} using supervised learning techniques or applying N-gram technique to categorize newspaper corpora~\cite{mansur2006analysis}, dataset preparation~\cite{rahman2018datasets} for aspect-based sentiment analysis~\cite{sumit2018exploring}, and word embeddings for document classification~\cite{ahmad2016bengali}. 

Apart from these, some approaches focus on parts of speech~(PoS) tagging techniques for parsing texts for rich semantic processing by combining Maximum Entropy, Conditional Random Field~(CRF) and SVM classifiers using weighted voting techniques~\cite{ekbal2009voted} or based on bidirectional LSTM—CRF based network~\cite{alam2016bidirectional}. 
In these approaches, classic ML algorithms such as SVM, KNN, NB, RF, and GBT have been used. These approaches, either lack comprehensive linguist analysis or are carried out on a small-scale labeled dataset with limited vocabulary, which diminishes the performance of supervised learning algorithms. 
IN contrast, a growing body of work aims to understand and detect hate speech by creating representations for content on the web and then classifying these tweets or comments as hateful or not, drawing insights along the way~\cite{hate1}. 
Many efforts have been made to classify hate speech using data scraped from online message forums and popular social media including Twitter, YouTube, Facebook, and Google, with their Perspective API~\cite{salminen2018anatomy}. 

A LR-based approach is proposed for one-to-four-character n-grams for classifying tweets that are labeled as racist, sexist or neither~\cite{waseem2016hateful} and for distinguishing hateful and offensive but not hateful tweets with \texttt{L2} regularization~\cite{alorainy2018cyber}.
For the latter use case, the authors applied word-level n-grams and various PoS, sentiment and tweet-level metadata features. 
However, the accuracy of these solutions are not satisfactory~\cite{islam2017comparative}. 
Davidson et al.~\cite{davidson2017automated} used DNN approaches with two binary classifiers: one to predict the presence of abusive speech more generally, and another one to discern the form of abusive speech or building a classifier composed of two separate networks for hate speech detection~\cite{kshirsagar2018predictive}.

Existing classifiers are predominantly supervised~\cite{wei2017convolution}: LR is the most popular, while other algorithms, e.g. SVM, NB, and RF are also used.
Despite diverse types of features introduced, little is known about the multimodality of different types of features in such a single classifier.
Most methods simply `use them all' by concatenating all feature types into high-dimensional, sparse feature vectors that are prone to overfitting, especially on short texts e.g. tweets. 
Some applied an automated statistical feature selection process to reduce and optimize the feature space, while others did this manually.

Since the impact of feature selection is unknown, whether different types of features are contributing to better classification accuracy remains questionable.
Hence, classic methods that have been investigated in the literature employ a  abundance of task-specific features. 
It is not clear what their individual contributions could be~\cite{fortuna2018survey}. 
These approaches in comparison with DL-based approaches is rather incomparable because of the efficiency of linear models at dealing with billions of such texts proven less accurate and unscalable, which is probably another primary reason. 
DL is applied to learn abstract feature representations used for the text classification. The input can take various forms of feature encoding, including any of those used in the classic methods. 
However, the key difference is that the input features are not directly used for classification. 
Instead, multilayer structures learn abstract representation  of the features, which is found more effective for learning, as typical manual feature engineering is then shifted to the network topology, which is carefully designed to automatically extract useful features from a simple input feature representation~\cite{fortuna2018survey}.

CNN, Gated Recurrent Unit~(GRU)~\cite{zhang2018detecting}, and LSTM networks are the most popular architectures in the literature. 
In particular, CNN is an effective arsenal to act as `feature extractors', whereas LSTM is a type of powerful recurrent type network for modeling orderly sequence learning problems. 
We also observe the use of LSTM or GRU with pre-trained \texttt{Word2Vec}, \texttt{GloVe}, and \texttt{fastText} embeddings to fed into a CNN with max pooling to produce input vectors for a neural network~\cite{zhang2018}, because, in the context of text
classification, CNN extracts word or character combinations, e.g., phrases, n-grams and LSTM learns a long-range word or character dependencies in texts. 
In theory, the \texttt{Conv-LSTM} is a powerful architecture to capture long-term dependencies between features extracted by CNN. In practice, they are found to be more effective than structures solely based on CNN or LSTM in tasks such as drug-drug interaction predictions~\cite{karim2019DDI}, network intrusion detection~\cite{khan2019scalable}, and gesture/activity recognition where the networks learn temporal evolution of different regions between frames, and Named Entity Recognition~(NER)~\cite{Conv_LSTM1} where the class of a word sequence can depend on the class of its preceding word sequence. 

While each type of network has shown effectiveness for general purpose text classification, few works have explored combining both structures into a single network~\cite{hate3}, except that using this combination, and especially when trained using transfer learning achieved higher classification accuracy than either neural network classifier alone~\cite{hate5}. 
In contrast, our approach relies on a word embedding similar to fastText but with a strong focus on the under-resourced setting by reducing the number of required parameters and length of training required, while still yielding improved performance and resilience across related classification tasks. 
We hypothesize that \texttt{MConv-LSTM} can also be effective not only for hate speech detection but also for sentiment analysis and document classification since this deep architecture can capture co-occurring word n-grams as useful patterns for classification~\cite{joulin2016fasttext,fastText}. 
Moreover, we believe that our network will be able to learn flexible vector representations that demonstrate associations between words typically used not only for hate speech detection but also for document classification and sentiment analysis.

\section{Datasets}
\label{section:3}
One major limitation of state-of-the-art approaches is the lack of comparative evaluation on publicly available data sets for the Bengali language. 
A large majority of the existing works are evaluated on privately collected datasets, often targeting different problems, and most of them contain only a small Bengali text corpus. 
A careful analysis was carried out while differentiating linguistic features, e.g. abusive language can be different from hate speech but is often used to express hate as shown in examples in~\cref{cdec_wf1}. 
To best of our knowledge, there are only a few works proposed on Bengali word embeddings. The Polyglot project~\cite{al2013polyglot} is one of the first attempts, which in which the word embeddings are trained for more than 100 languages using their corresponding Wikipedia dumps in which only 55K corpus from the Bengali Wikipedia dump were used to train the model to generate the embeddings. 
The second approach~\cite{ahmad2016bengali} built an embedding model on 200K Bengali corpus from 13 news articles. 
Third, a neural lemmatizer is proposed by \cite{chakrabarty2016benlem} using Bengali word embeddings, which is built on even a smaller dataset of 20K corpus.

The fourth one called fastText~\cite{fastText}, which is trained on common crawl and Wikipedia using CBOW with position-weights, in dimension 300, with character n-grams of length 5, a window of size 5 and 10 negatives. Eventually, fastText distributes the pre-trained word vectors for 157 languages. 
To further explore and apply more extensive linguist analysis on Bengali with flexibly and efficiently, we not only collected the largest corpus of Bengali raw texts from Wikipedia but also prepared three more datasets: documents of contemporary Bengali topics, texts expressing hate speech~(political, religious, personal, geopolitical and gender abusive) and public sentiments~(about Cricket, products, movie, and politics).  

\subsection{Raw text corpus collection}
For our corpus, Bengali articles were collected from numerous sources from Bangladesh and India including a Bengali Wikipedia dump, Bengali news articles~(Daily Prothom Alo, Daily Jugontor, Daily Nayadiganta, Anandabazar Patrika, Dainik Jugasankha, BBC, and Deutsche Welle), news dumps of TV channels~(NTV, ETV Bangla, ZEE News), books, blogs, sports portal, and social media (Twitter, Facebook pages and groups, LinkedIn). We also categorized the raw articles for ease preprocessing for later stage. 
Facebook pages~(e.g. celebrities, athletes, sports, and politicians) and newspaper sources were scrutinized because composedly they have about 50 million followers and many opinions, hate speech, and review texts come or spread out from there. 
All text samples were collected from publicly available sources. 
Altogether, our raw text corpus consists of 250 million articles.
Then two linguists and three native Bengali speakers annotated the samples from separate datasets for hate speech detection, document classification, and sentiment analysis, which is described in subsequent subsections\footnote{The prepared dataset, embedding model, and codes will be made publicly available.} to encourage future comparative evaluation. 

\begin{figure*}[h]
	\centering
	\begin{subfigure}{0.48\linewidth}
		\centering
		\includegraphics[width=\linewidth]{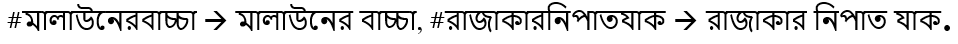}
		\caption{Hashtags normalization}
        \label{fig:norm}
	\end{subfigure}
	\begin{subfigure}{0.48\linewidth}
		\centering
		\includegraphics[width=0.65\linewidth]{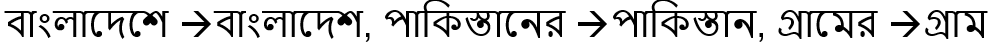}
		\caption{Stemming}
        \label{fig:stem}
	\end{subfigure}
	\caption{Example of hashtags normalization and stemming} 
	\label{fig:norm_stem}
\end{figure*}

\Cref{clouds} shows the frequency word clouds of potential creative hates expressed in our datasets. Since the collected Bengali text corpus comes with various dimensional, sparse and relatively noisy, before the word embedding step and manual labeling, emoticon, digits~(Bengali and English) special characters, hash-tags and excess use of spaces~(due to redundant use of spaces, tabs, shift) were removed to give annotators an unbiased-text-only content to make a decision based on criteria of the objective. 

\subsection{Data preprocessing}
\label{data_prep}
Since the obtained corpus is rather noisy, we perform an automated cleaning for all datasets before giving it to the annotators.
First, we remove HTML markup, links, and image titles. 
From text, digits, special characters, hash-tags and excessive white space are removed.
Then the following further steps are performed:

\vspace{1mm}
\begin{itemize}
    \item \textbf{PoS tagging:} an BLSTM-CRF based approach suggested in literature \cite{alam2016bidirectional} was applied.
    \item \textbf{Removal of proper nouns:} various proper nouns and noun suffixes were identified and replaced with tags to provide ambiguity. 
    \item \textbf{Hashtags normalization:} hastags in tweets, comments, and posts were normalized using a dictionary-based look up\footnote{Based on \url{https://www.101languages.net/bengali/bengali-word-list/}}, which are often used to compose words (e.g. \cref{fig:norm}) was used to split such hashtags. 
    \item \textbf{Stemming:} inflected words are reduced to their stem, base or root form in order to reduce dimension and obtain more standard Bengali tokens (e.g. \cref{fig:stem})(English translation: Bangladeshe to Bangladesh, Pakistaner to Pakistan, from village to village).
    \item \textbf{Stop word removal:} commonly used Bengali stop words~(provided by \texttt{Stopwords ISO}\footnote{\url{https://github.com/stopwords-iso/stopwords-bn}}) are removed.
    \item \textbf{Removing infrequent words:} finally, we remove any tokens with a document frequency less than 5. 
\end{itemize}
\vspace{1mm}

\subsection{Dataset for document classification}
Despite several comprehensive textual datasets available for document categorization, only a few labeled datasets exist for Bengali articles. Sophisticated supervised learning approaches, therefore, were not a viable option to address document classification problems. Eventhough, large-scale corpora for other language are available online, it is rare for under-resourced languages like Bengali. Hence we decided to build our own corpus. As described above, the articles are collected into different categories such that each belongs to any one of the following categories: state, economy, entertainment, international, and sports as outlined in \cref{table:doc}. 

\begin{table}
    \centering
    \caption{Statistics of document classification dataset. For each category, we show the number of documents, number of words, average number of sentences per document, and average number of words per sentence}
    \label{table:doc}
    \vspace{-3mm}
    \scriptsize{
    \begin{tabular}{lllll}
        \hline \hline
        \textbf{Category}  & \textbf{\#Doc}  & \textbf{\#Words}  &	\textbf{ASD} &  \textbf{AWS}\\
        \hline
        State &	242,860 &	57,019,465 &	18.50 &	13.356 \\
        \hline
        Economy &	18,982 &	4,915,141 &	20.18 &	13.378 \\
        \hline
        International &	32,203 &	7,096,111 &	18.47 &	12.493 \\
        \hline
        Entertainment &	31,293 &	6,706,563 &	21.70 &	10.236 \\
        \hline
        Sports &	50,888 &	12,397,415 &	22.80 &	11.069 \\
        \hline
        \textbf{Total} &	376,226 &	88,134,695 &	20.33 &	12.10 \\
        \hline
    \end{tabular}
    }
\end{table}

\subsection{Hate speech dataset preparation} 
To prepare the data we used a bootstrapping approach, which starts with an initial search for specific types of texts, articles or tweets containing common slurs and terms used pertaining to targeting characteristics.
We manually identify frequently occurring terms in the texts containing hate speech and references to specific entities.  
From the raw texts, we further annotated 100,000 statements, texts or articles, which directly or indirectly express `hate speech'.

\begin{table}
    \centering
    \caption{Statistics of the hate speech detection dataset}
    \label{table:hate}
    \vspace{-3mm}
    \scriptsize{
    \begin{tabular}{llll}
        \hline \hline
        \textbf{Type of hate} &  \textbf{\#Statement}  & \textbf{\#Words}  & \textbf{AWS}\\
        \hline
        Political &	7,182 &	132,867 &	18.50 \\
        \hline
        Religious &	6,975 &	140,756 &	20.18  \\
        \hline
        Gender abusive &	7,300 &	134,831 &	18.47  \\
        \hline
        Geopolitical &	6,793 &	117,180 &	17.25  \\
        \hline
        Personal &	6,750 &	146,475 &	21.70  \\
        \hline
        \textbf{Total}  &	35,000 &	672,109 &	19.22  \\
        \hline
    \end{tabular}
    }
\end{table}

However, we encountered numerous challenges e.g. since there exists different types of hate in the regions, distinguishing hate speech from non-hate offensive language is a challenging task, which is because sometimes they overlap but are not the same~\cite{zhang2018hate}. 
Collected data samples are annotated in a  semi-automated way (after removing mostly similar statements). 
First, we attempted to annotate them in an automated way, which is inspired by the comparing communities technique using \texttt{creative
political slang~(CSPlang)} \cite{political_slang}. 
We prepare normalized frequency vectors of 175 abusive terms\footnote{\url{https://github.com/rezacsedu/BengFastText}} across the raw texts, which are commonly used to express hates in Bengali language. 
Then we assign label `hate' if at least one of these terms exists in the text, otherwise, we put label `neutral'. This way, however, differentiating 'neutral' and `non-hate' was not possible. 

\begin{figure*}
	\centering
	\begin{subfigure}{.5\textwidth}
		\centering
		\includegraphics[width=0.7\linewidth,height=35mm]{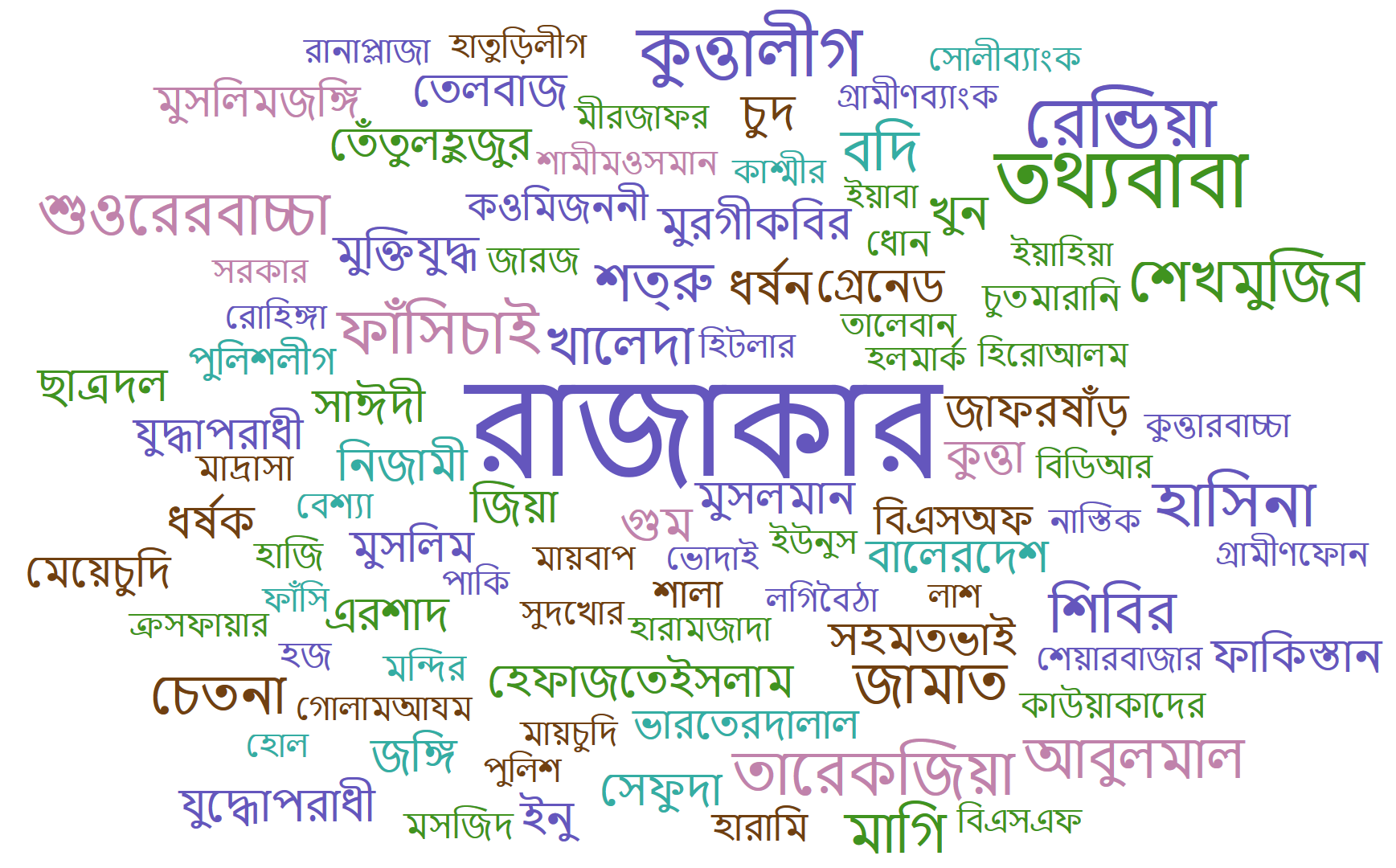}
		\caption{Frequently used terms in hate speech}
        \label{clouds:hate}
	\end{subfigure}%
	\begin{subfigure}{.5\textwidth}
		\centering
		\includegraphics[width=0.7\linewidth,height=35mm]{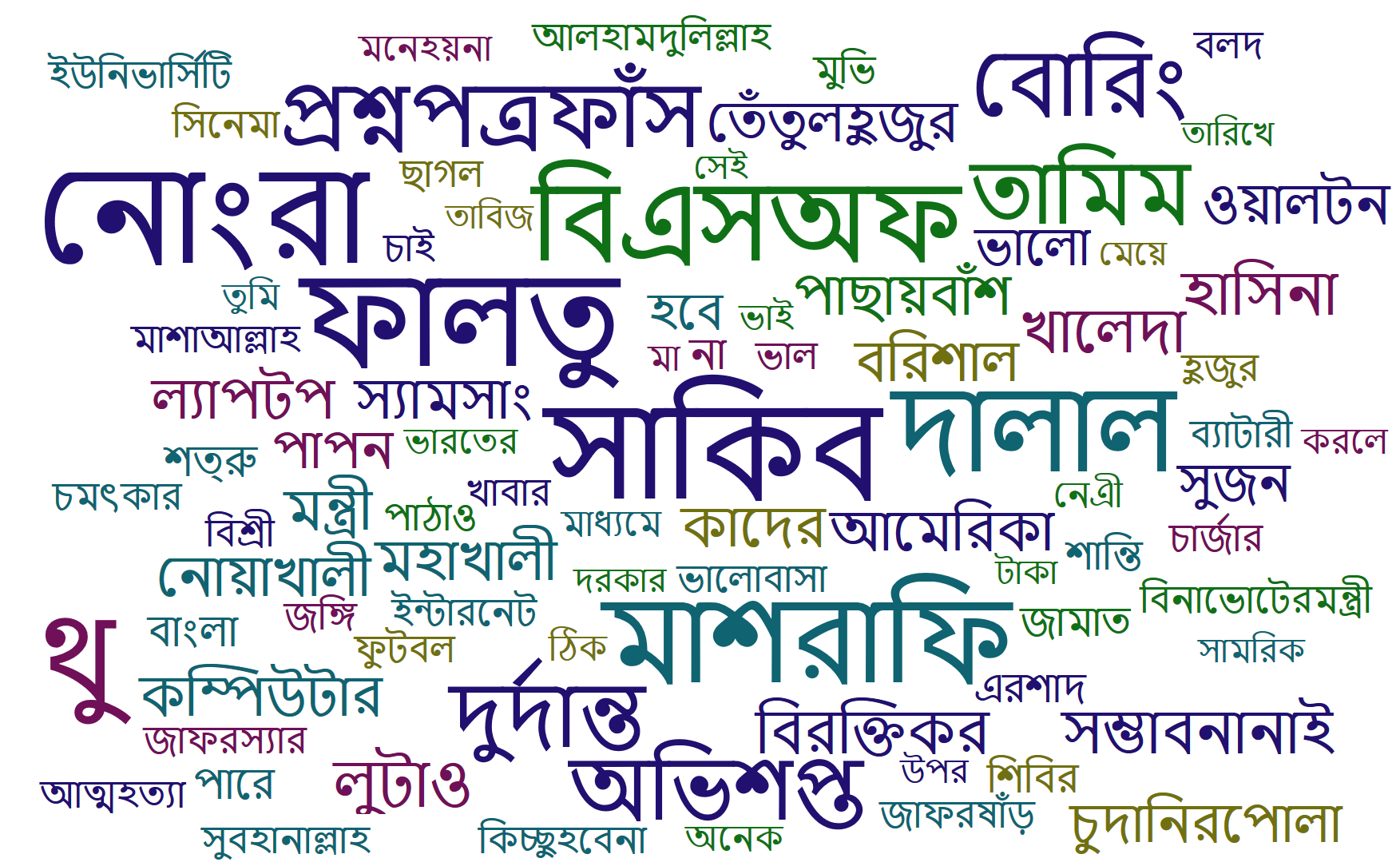}
		\caption{Frequently used terms in sentiments}
        \label{clouds:sentiment}
	\end{subfigure}
		\begin{subfigure}{.5\textwidth}
		\centering
		\includegraphics[width=0.7\linewidth,height=40mm]{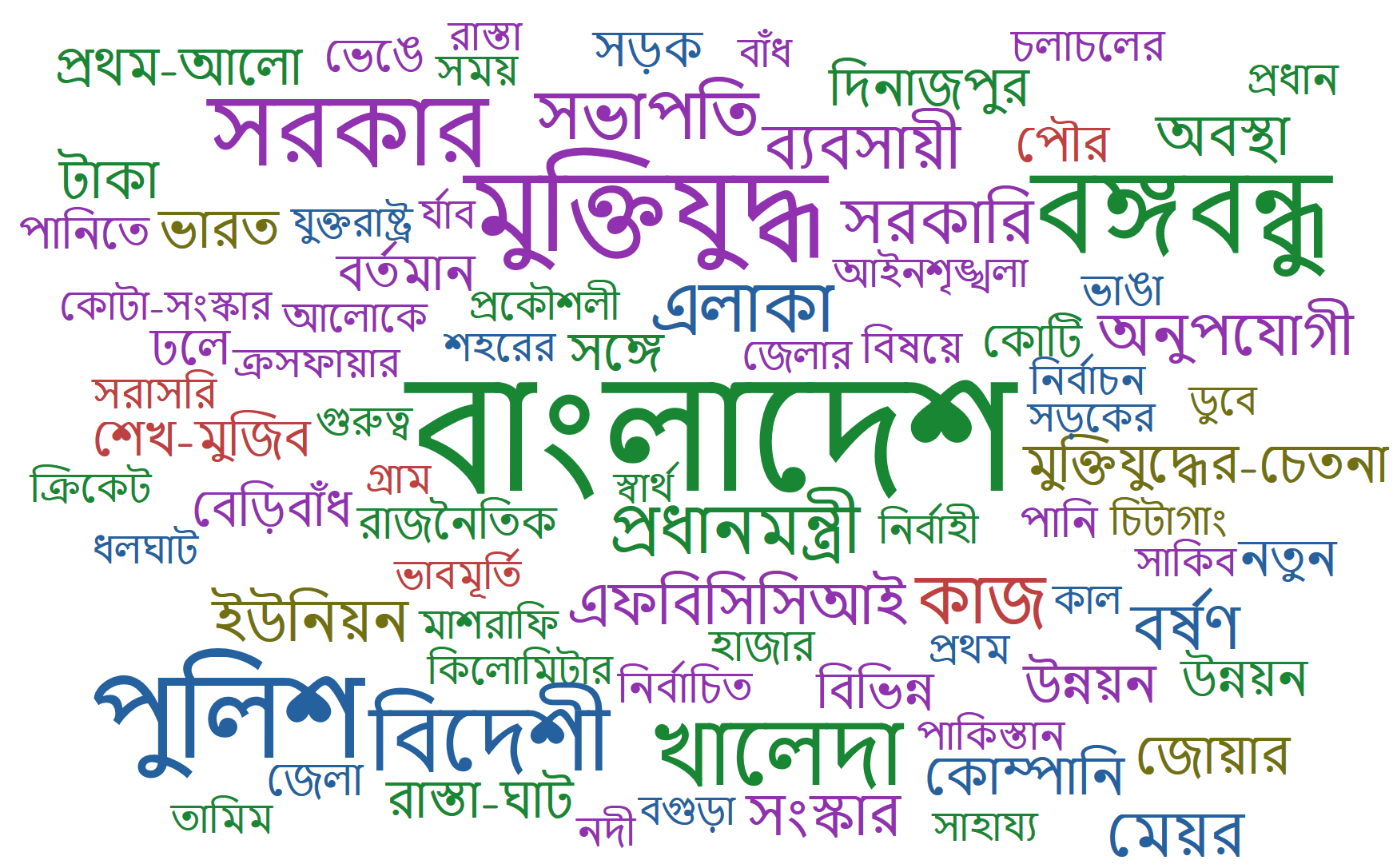}
		\caption{Frequently used topics of discussion}
        \label{clouds:topics}
	\end{subfigure}
	\caption{Frequency word clouds on potential hates, sentiments, and common topics}
	\label{clouds}
\end{figure*}
\raggedbottom

The annotations are further validated and corrected by three experts (one South Asian linguist and two native speakers) into one of three categories: hate, non-hate and neutral. 
To reduce possible bias, each label was assigned based on a majority voting on the annotator's independent opinions.
Fortunately, certain types of hate were easy to identify and 
annotate based on \texttt{CSPlang} that are nonstandard word for Bengali, which conveys a positive or negative attitude towards a person, a group of people, or an issue that is the subject of discussion in political discourse \cite{political_slang} and can be easily annotated as a political hates. 
Finally, non-hate and neutral statements were removed from the list and hates were further categorized into political, personal, gender abusive, geopolitical, and religious hates. 
We found that from the annotated text 3.5\% was classified as hate speech, 
which resulted in 35,000 statements labeled as hate. 
\Cref{clouds:hate} shows the most frequently used terms expressing hates; whereas, some statistics can be found in \cref{table:hate}.

\subsection{Dataset for sentiment analysis}
Our empirical study found that opinions, reviews, comment or reactions about news or someone's status updates are usually provided in Bengali. 
However, research has observed that about 20\% of express sentiments in English \cite{rahman2018datasets}.
Also, some comments were only emoticons or written in non-native scripts.
As English and emoticons are already investigated elsewhere, we consider this outside of the scope of the current work.

\begin{table}[h]
    \centering
    \caption{Statistics of the dataset for sentiment analysis}
    \label{table:sentiment}
    \vspace{-3mm}
    \scriptsize{
    \begin{tabular}{llll}
        \hline \hline
        \textbf{Polarity} &	\textbf{\#Statement}  & \textbf{\#Words}   &  \textbf{AWS} \\
        \hline
        Negative &	153,000  &	727,3200  &	40.16 \\
        \hline
        Positive &	167,000 &	557,3550  &	30.65  \\
        \hline
        \textbf{Total} &	320,000 &	12,846750  &	35.40  \\
        \hline
    \end{tabular}
    }
\end{table}

The annotation was done manually with the help of native Bengali speakers and majority voting~(with ambiguous polarity removed), which left 320,000 reviews categorized either positive or negative. 
\Cref{clouds:hate} shows the most frequently used terms expressing hate; whereas, related statistics can be found in~\cref{table:sentiment}. 

Finally, we follow a similar preprocessing steps for extracting most important words, which is a statement-based observation that states that in a collection of data, the frequency of a given word should be inversely proportional to its rank in the corpus --that is most frequent word scores rank 1 in a dataset should occur approximately twice as the second most frequent word and three times as the third most frequent, and so on.

\section{Methods}
\label{nettwork}
To validate the usefulness of the dataset and the  \texttt{BengFastText} model, we perform several ML tasks, which are discussed including word embeddings, network construction, training, and hyper-parameter optimization.

\subsection{Neural word embeddings}
Given a document, e.g., an article, a tweet, or a review of Bengali text, we apply preprocessing to normalize the text~(see \cref{data_prep}). 
The preprocess reduces vocabulary size in the dataset due to the colloquial nature of the texts and to some degree, addresses the sparsity in word-based feature representations. 
We also tested by keeping word inflections, using lemmatization instead of stemming, and lower document frequencies. 
Empirically we found that using lemmatization to contribute to slightly better accuracy, hence reported here subsequently.
Each token from the preprocessed input is embedded into a 300-dimensional real-valued vector, where each element is the weight for the dimension for the token. For later classification tasks, we constrain each sequence to 100 words, truncating long texts and pad shorter ones with zero values. 

However, one issue with this design specific choice forces us to consider 300 words per document for document classification otherwise for the majority of the documents inputs to the convolutional layers will be padded with many blank vectors, which causes the network to perform poorly. 
First, the Word2Vec model is built based on skip-gram method since it is computationally faster and efficient than CBOW for large-scale text corpus~\cite{bian2014knowledge} and used for Bengali sentiment analysis~\cite{sumit2018exploring}. 
From a given a sequence of words $(w_1,w_2,...,w_n)\in \mathcal{C}$, the neural network model aims to maximize the average log probability $L_p$ (see \cref{eq:log}) according to the context within the fixed-size window, in which c represents a context~\cite{mikolov2013efficient}. 

\begin{equation}
    L_p= \frac{1}{N} \sum_{n=1}^{N} \sum_{-c \leq j \leq c, j \neq 0} \log p\left(w_{n+j} | w_{n}\right)
    \label{eq:log}
\end{equation}
\vspace{-2mm} 
\begin{equation}
    \log \sigma\left(v_{w_{O}}^{\prime \top} v_{w_{I}}\right)+\sum_{i=1}^{k} \mathbb{E}_{w_{i}} \sim_{P_{n}(w)}\left[\log \sigma\left(-v_{w_{i}}^{\prime \top} v_{w_{I}}\right)\right]
    \label{eq:kutta}
\end{equation}

To define $p\left(w_{n+j} | w_{n}\right)$, we use negative sampling by replacing $\log p\left(w_{O} | w_{I}\right)$ with a function to discriminate target words $(w_o)$ from a noise distribution $P_n(w)$ drawing $k$ words from $P_n(w)$ in \cref{eq:kutta}. 
Eventually, the embedding of a word $s$ occurring in corpus $\mathcal{C}$ is the vector $v_s$ in~\cref{eq:kutta} derived by maximizing~\cref{eq:log}. 

\begin{figure*}
	\centering
	\includegraphics[width=0.7\textwidth]{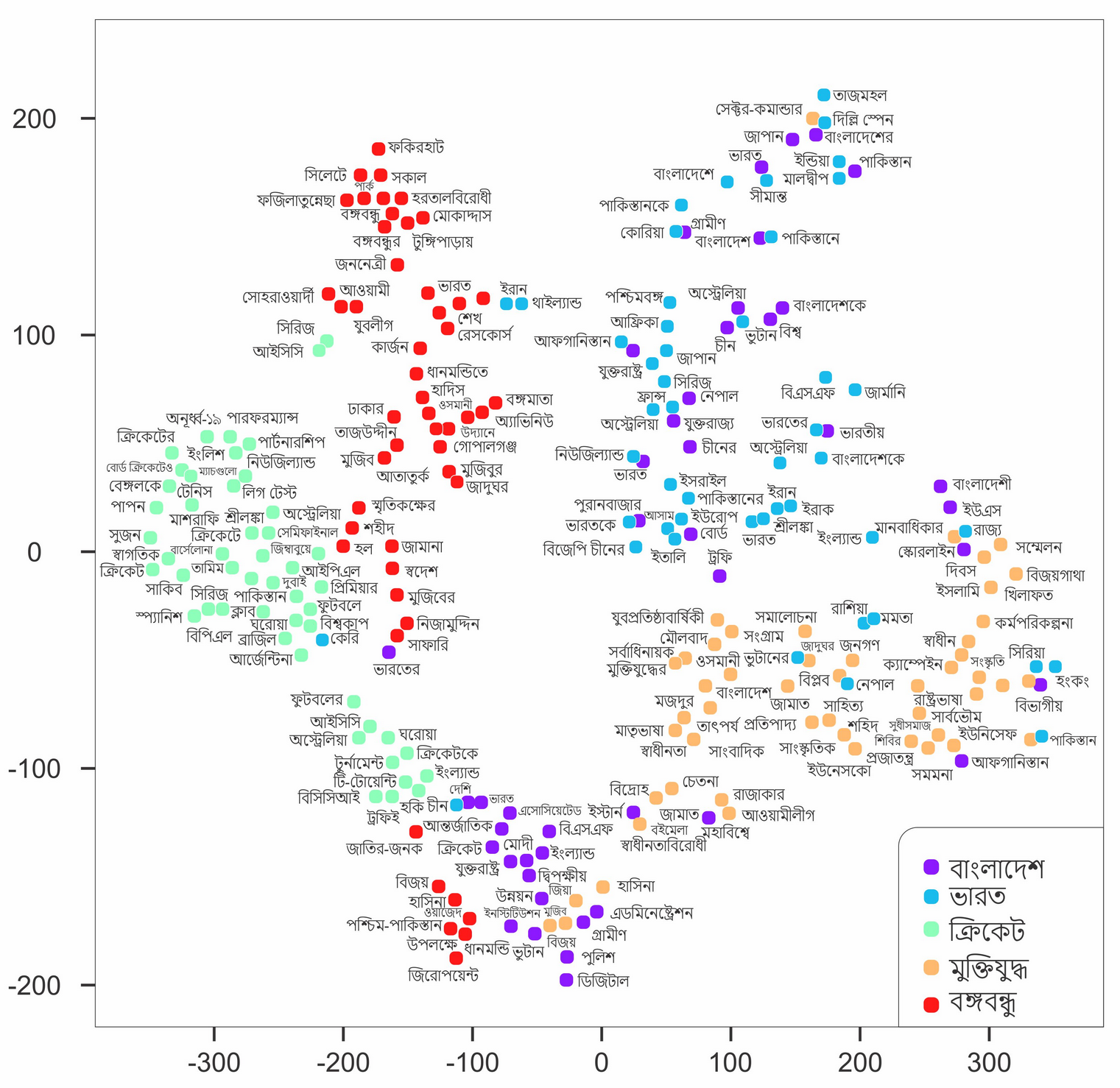}	
    \caption{t-SNE representation of the embeddings, where closer words are semantically and statistically more related}	
	\label{fig:semantic_similarity}
\end{figure*}

Then we built \texttt{BengFastText} model based on GloVe for creating the embeddings~\cite{GloVe}. In which, instead of extracting the embeddings from a neural network that is designed to perform a different task like predicting neighboring words~ (i.e., CBOW) or predicting the focus word~(i.e., skip-gram), the embeddings are optimized directly, so that the dot product of two word vectors equals the log of the number of times the two words will occur near each other.

Finally, we employed fastText for creating the embeddings~\cite{joulin2016fasttext} by the \texttt{BengFastText} model. 
Instead of learning vectors for words directly, fastText represents each word as an n-gram of characters, which helps capture the meaning of shorter words and allows the embeddings to understand suffixes and prefixes. Once the word is represented using character n-grams, a skip-gram model with negative sampling based on \cite{word2vec} is used to score similarity between a fixed target word $w_{t}$ and a word within the context $w_{c}$ to learn the embeddings. 
This approach intuitively sounds similar to Word2Vec and can be considered to be a CBOW model with a sliding window over a word because no internal structure of the word is taken into account. As long as the characters are within this window, the order of the n-grams does not matter. The difference, however, is that the similarity function is not a direct dot product between the word vectors, rather it is a dot product between two words represented by sums of n-gram vectors. Then a latent text embedding is derived as an average of the word embeddings, and the text embedding is used to predict the label. The overall objective function is to minimize the following loss function~\cite{joulin2016fasttext,fastText}: 

\begin{equation}
-\frac{1}{N} \sum_{n=1}^{N} y_{n} \log \left(f\left(B A x_{n}\right)\right)
\label{eq:fastText}
\end{equation}

where $x_{n}$ represents an n-gram feature, A represents the lookup table to an average text embedding, and B converts the embedding to pre-softmax values for each class, while a hierarchical softmax~\cite{hierarchicalsoftmax} is applied for the text classification~\cite{joulin2016bag} given the large number of classes to minimize computational complexity. 

\subsection{Network construction}
In CNN, convolutional filters are used to capture the local dependencies between neighbor words. However, a fixed filter lengths makes CNN model hard to learn overall dependencies of a whole sentence. Following steps are employed to overcome this limitation: 

\vspace{1mm}
\begin{itemize}
  \item We consider 3-channel CNN approach by setting a fixed filter size with different sized kernels. This allows a text to be processed at different n-grams at a time. The concatenated vector helps the model to learn how to best integrate these interpretations altogether, which can be considered as a representation vector constructed from every channel concatenating local relationship values.

  \item Then employ an LSTM layer, which is because an LSTM can preserve long-term dependency among features. Hence, a feature vector constructed by LSTM can carry overall dependencies of a whole sentence more efficiently~\cite{karim2019DDI}.  

  \item We concatenate both multichannel-CNN and LSTM layers to get a shared feature representation: CNN layers capture local relationship, LSTM layers carry overall relationship). 
\end{itemize}
\vspace{1mm}

We expect the shared representation layer containing two vectors to model the non-linearity more efficiently for our tasks, e.g., hate speech detection, document classification, or sentiment analysis. Given a text expressing hates~(hopefully), the \texttt{MConv-LSTM} model either reuse the weight matrix from the pre-trained \texttt{BengFastText} model or generates a word embedding representation by an embedding layer. 
This means that we utilize two types of word embeddings to train the models: i) we initialize the weights in the embedding layer randomly and let our model learn the embeddings on the fly, ii) we use pre-trained word embeddings from the GloVe, Word2Vec, and \texttt{BengFastText} models to set the weights of our embedding layer as a custom embedding layer. 

\begin{figure*}
	\centering
	\includegraphics[width=\textwidth,height=75mm]{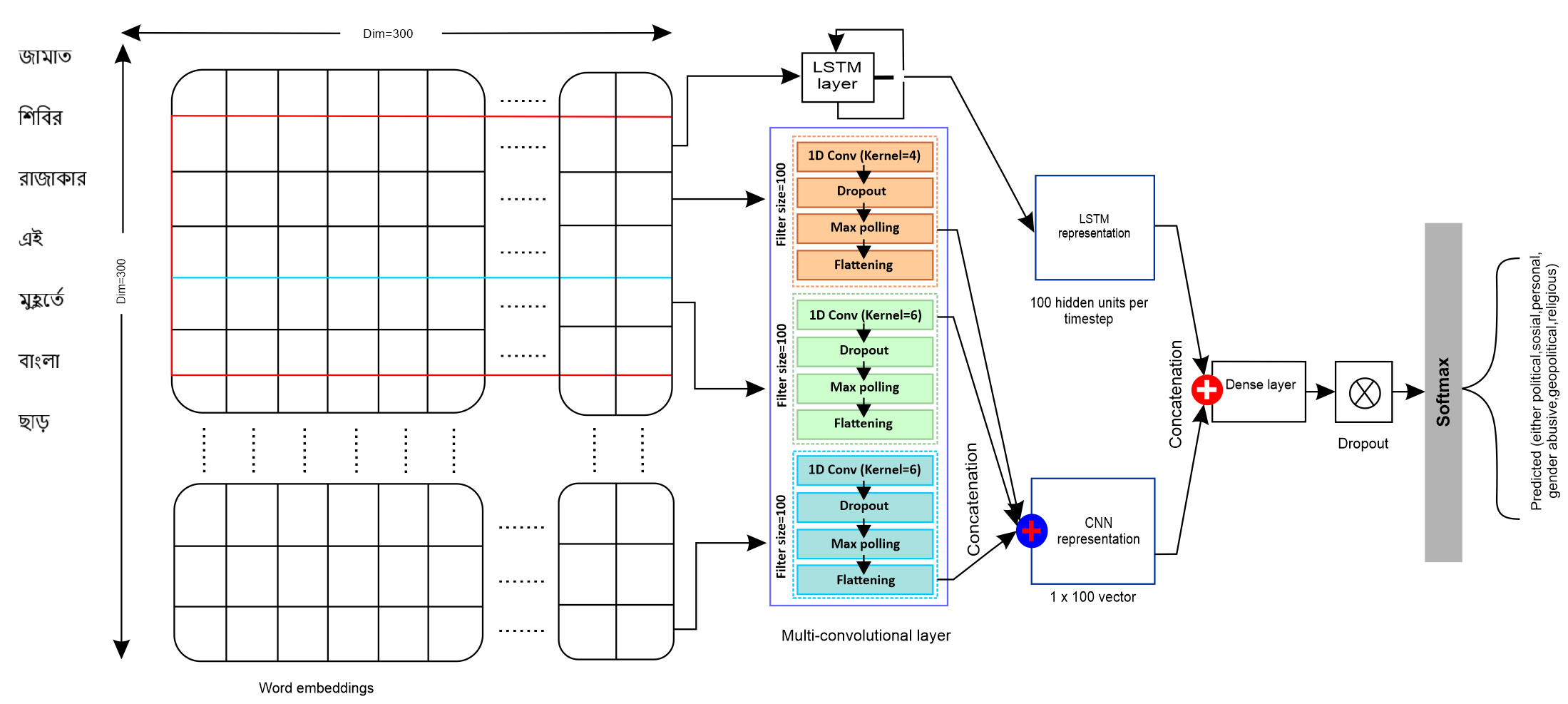}	
    \caption{A schematic representation of the \texttt{MConv-LSTM} network, which starts from taking an input into a n-dimensional embedding hyperspace and passing to both CNN and LSTM layers before getting a concatenated vector representation to fed through dense, dropout, and Softmax layers for predicting the hate class}	
	\label{fig:network_architecture}
\end{figure*}

Using either option, we construct and train the \texttt{MConv-LSTM} network architecture shown in \cref{fig:network_architecture} in which the rectified linear unit~(ReLU) is used as the activation function for the hidden layers and Softmax in the output layer. Once the word embedding layer maps each text as a `sequence' into a real vector domain, the embedding representation is fed into both LSTM and multi-CNN layers towards getting a shared feature representation. 
To capture both the local and global relationships, we extended the \texttt{Conv-LSTM} network proposed in literature~\cite{Conv_LSTM1} in which each input $\mathcal{X}_{1},\mathcal{X}_{2},...,\mathcal{X}_{t}$, cell outputs $\mathcal{C}_{1},\mathcal{C}_{2},...,\mathcal{C}_{t}$, hidden states $\mathcal{H}_{1},\mathcal{H}_{2}....,\mathcal{H}_{t}$, and gates $i_t$,$f_t$,$o_t$ of the network~\cite{zhang2018hate}.
\texttt{Conv-LSTM} determines the future state of a certain cell in the input hyperspace by the inputs and past states of its local neighbors, which is achieved by using a conv operator in the state-to-state and input-to-state transitions as~\cite{Conv_LSTM1}:

\begin{align} 
    i_{t} &=\sigma\left(W_{x i} * \mathcal{X}_{t}+W_{h i} * \mathcal{H}_{t-1}+W_{c i} \circ \mathcal{C}_{t-1}+b_{i}\right) \\ 
    f_{t} &=\sigma\left(W_{x f} * \mathcal{X}_{t}+W_{h f} * \mathcal{H}_{t-1}+W_{c f} \circ \mathcal{C}_{t-1}+b_{f}\right) \\ \mathcal{C}_{t} &=f_{t} \circ \mathcal{C}_{t-1}+i_{t} \circ \tanh \left(W_{x c} * \mathcal{X}_{t}+W_{h c} * \mathcal{H}_{t-1}+b_{c}\right) \\
    o_{t} &=\sigma\left(W_{x o} * \mathcal{X}_{t}+W_{h o} * \mathcal{H}_{t-1}+W_{c o} \circ \mathcal{C}_{t}+b_{o}\right) \\ \mathcal{H}_{t} &=o_{t} \circ \tanh \left(\mathcal{C}_{t}\right) 
\end{align} 
\vspace{1mm}

where `*' denotes the conv operator and `o' is the entrywise multiplication of two matrices of the same dimension. The second LSTM layer emits an output `H', which is then reshaped into a feature sequence and feed into fully connected layers to make the prediction at the next step and as an input at the next time step. Intuitively, an LSTM layer treats an input feature space of 100$\times$300 and it's embedded feature vector's dimension as timesteps and outputs 100 hidden units per timestep. Once the embedding layer passes an input feature space 100$\times$300 into three channels~(i.e. a multi-1D convolutional layer) with 100 filters each but with different kernel-size of 4, 6 and 8, respectively (i.e. 4-grams, 6-grams, and 8-grams of hate speech text). 

We pad the input such that the output has the same length as the original input. Then the output of each convolutional layer is pass to three separate dropout layers to regularize learning to avoid overfitting \cite{srivastava2014dropout}. Intuitively, this can be thought of as randomly removing a word in sentences and forcing the classification not to rely on any individual words. This involves the input feature space into a 100$\times$100 representation, which is then further down-sampled by three different 1D max pooling layers, each having a pool size of 4 along the word dimension, each producing an output of shape 25$\times$100. Where each of the 25 dimensions can be considered as an`extracted feature'. 
Each max pooling layer follows to`flatten' the output space by taking the highest value in each timestep dimension. This produces a 1$\times$100 vector, which forces words that are highly indicative of interest. Then these vectors as two information channels (from CNN + LSTM) are then concatenated and fed into the neural network after passing through another dropout layer. Finally, the concatenated representation is passed to a fully connected Softmax to predict a probability distribution over the classes for the specific tasks.

\subsection{Network training}
The first-order gradient-based AdaGrad optimizer is used to learn model parameters with varying learning rates and a batch size of 128 to optimize the cross-entropy loss~(\texttt{categorical cross-entropy} in case of document classification and hate speech detection and \texttt{binary cross-entropy} loss in case of sentiment analysis) of the predicted label of the text P vs the actual label of the text T.

\begin{align} 
L = \sum_{i, j, k} T_{i, j, k} \log P_{i, j, k}+\left(1-T_{i, j, k}\right) \log \left(1-P_{i, j, k}\right)
\label{eq:cross_entropy}
\end{align} 

We perform hyperparameters optimization with random search and 5-fold cross-validation. Further, we observed the performance by adding Gaussian noise layers, followed by Conv and LSTM layers to improve model generalization and reduce overfitting. 
Out of vocabulary~(OOV), particularly in word embedding models, is a known issue. The OOV cannot be tackled by existing language models like Polyglot, Word2Vec, and GloVe.
This potential issue with the pre-trained Word2Vec model were addressed based on \cite{wei2017convolution}: i) randomly setting the OOV word vector following a continuous uniform distribution, ii) randomly selecting an in-vocabulary word from the pretrained model to use its vector. 

For the latter, surrounding-contexts are considered, given that a low-magnitude vector may not be appropriate at the start of training~\cite{bojanowski2017enriching}. 
Both settings are found to be empirically helpful in our experiments. However, the latter outperformed the former.
Other options could have been employed to get rid of OOV e.g. morphological classes~\cite{muller2011improved} or character-word LSTM language models~\cite{verwimp2017character} but are particularly applied to non-Bengali languages such as English and Dutch languages. 

\section{Experiments}
\label{er}
In this section, we discuss the results of the experiments based on and without the pre-trained \texttt{BengFastText} model. In addition, we provide a comparative analysis between \texttt{BengFastText} and baseline embedding methods: Word2Vec and GloVe. 

\subsection{Experiment setup}
All the methods were implemented in Python\footnote{\url{https://github.com/rezacsedu/BengFastText}} and experiments are carried out on a machine having a core i7 CPU, 32GB of RAM and Ubuntu 16.04 OS. 
The software stack consisting of Scikit-learn and Keras with the TensorFlow GPU backend, while the network training is performed on a Nvidia GTX 1050 GPU with CUDA and cuDNN enabled to make the overall pipeline faster. Open source implementations of Word2Vec\footnote{\textbf{Word2Vec}: \url{https://radimrehurek.com/gensim/models/word2vec.html}},  fastText\footnote{\textbf{fastText}: \url{https://github.com/Kyubyong/wordvectors}}, and GloVe\footnote{\textbf{GloVe}: \url{http://nlp.stanford.edu/projects/glove/}} are used to create the embeddings. For each experiment, 80\% of the data is used for the training with 5-fold cross-validation and testing the trained model on 20\% held-out data. 
Results based on best hyperparameters produced through empirical random search are reported here~(results based on embedding model are marked with + in subsequent tables). 

We trained LR, SVM, KNN, NB, RF, and GBT as baseline models using character n-grams and word uni-grams with TF-IDF weighting. Additionally, we report the results based on Word2Vec and GloVe embeddings methods. We report our results using macro-averaged precision, recall, and F1-score since the dataset is imbalanced. In case of hate speech detection, standard precision, recall, and F1-score are used. For the sentiment analysis, Matthias Correlation Coefficient~(\texttt{MCC}) and AUC score are measured the performance of the classifier. Finally, we perform model averaging ensemble~(MAE) of the top-3 models to report the final prediction. 

\begin{table*}[htp!]
    \caption{Document classification performance based on embedding methods}
    \label{table:all}
    \vspace{-3mm}
    \scriptsize{
    \begin{tabular}{c|l|l|l|l}
        \hline
        \textbf{Embedding method} & \textbf{Classifier} & \textbf{Precision} & \textbf{Recall} & \textbf{F1}\\ \hline
         \multirow{7}{*}{\textbf{BengFastText}} & LR & 0.754+3.8\% &	0.763+3.9\% &	0.756+3.85\% \\
         & NB & 0.734-\color{red}{2.11\%} &	0.736-\color{red}{2.10\%} &	0.737-\color{red}{2.0\%} \\
         & SVM & 0.766+3.9\% &	0.737+3.75\% &	0.739+3.65\% \\
         & KNN & 0.741+2.85\% &	0.745+2.75\% &	0.748+2.55\% \\
         & GBT & 0.825+5.65\% &	0.825+5.91\% &	0.826+5.85\%\\
         & RF & 0.813+5.7\% &	0.815+6.3\% &	0.822+6.24\%\\
         & MConv-LSTM  & 0.871+\color{blue}{8.75\%} &	0.883+\color{blue}{8.62\%} &	0.871+\color{blue}{8.75\%} \\
         & MAE & 0.883+\color{blue}{9.45\%} &	0.886+\color{blue}{9.25\%} &	0.892+\color{blue}{9.45\%} \\ \hline
         \multirow{7}{*}{\textbf{GloVe}} & LR & 0.725+1.7\% &	0.731+1.5\% &	0.723+1.7\% \\
         & NB & 0.692-\color{red}{3.25\%} &	0.701-\color{red}{3.1\%} &	0.693-\color{red}{3.2\%} \\
         & SVM & 0.724+2.2\% &	0.724+2.2\% &	0.718+2.4\% \\
         & KNN & 0.715+2.75\% &	0.716+2.54\% &	0.718+2.45\% \\
         & GBT & 0.772+2.32\% &	0.785+2.45\% &	0.779+2.65\%\\
         & RF & 0.792+4.15\% &	0.792+4.65\% &	0.795+4.75\%\\
         & MConv-LSTM  & 0.834+\color{blue}{6.4\%} &	0.846+\color{blue}{7.12\%} &	0.846+\color{blue}{7.31\%} \\
         & MAE & 0.859+\color{blue}{7.4\%} &	0.867+\color{blue}{7.2\%} &	0.864+\color{blue}{7.5\%} \\ \hline
        \multirow{7}{*}{\textbf{Word2Vec}} & LR & 0.746+2.4\% &	0.752+2.6\% &	0.749+2.7\% \\
         & NB & 0.721-\color{red}{2.15\%} &	0.717-\color{red}{2.1\%} &	0.719-\color{red}{2.2\%} \\
         & SVM & 0.741+2.7\% &	0.732+2.5\% &	0.736+2.6\% \\
         & KNN & 0.732+3.4\% &	0.725+3.6\% &	0.728+3.5\% \\
         & GBT & 0.806+4.854\% &	0.815+4.9\% &	0.810+5.1\%\\
         & RF & 0.792+5.6\% &	0.781+5.4\% &	0.786+5.5\%\\
         & MConv-LSTM  & 0.851+\color{blue}{7.9\%} &	0.862+\color{blue}{8.0\%} &	0.856+\color{blue}{8.1\%} \\
         & MAE & 0.865+\color{blue}{8.5\%} & 0.876+\color{blue}{9.1\%} &	0.870+\color{blue}{8.8\%} \\ \hline
    \end{tabular}
    }
\end{table*}

\subsection{\textbf{Analysis of document classification}} 
\label{sec:performance-document-classifier}
\begin{figure*}[htp!]
	\centering
	\begin{subfigure}{0.48\linewidth}
		\centering
		\includegraphics[width=\linewidth,height=75mm]{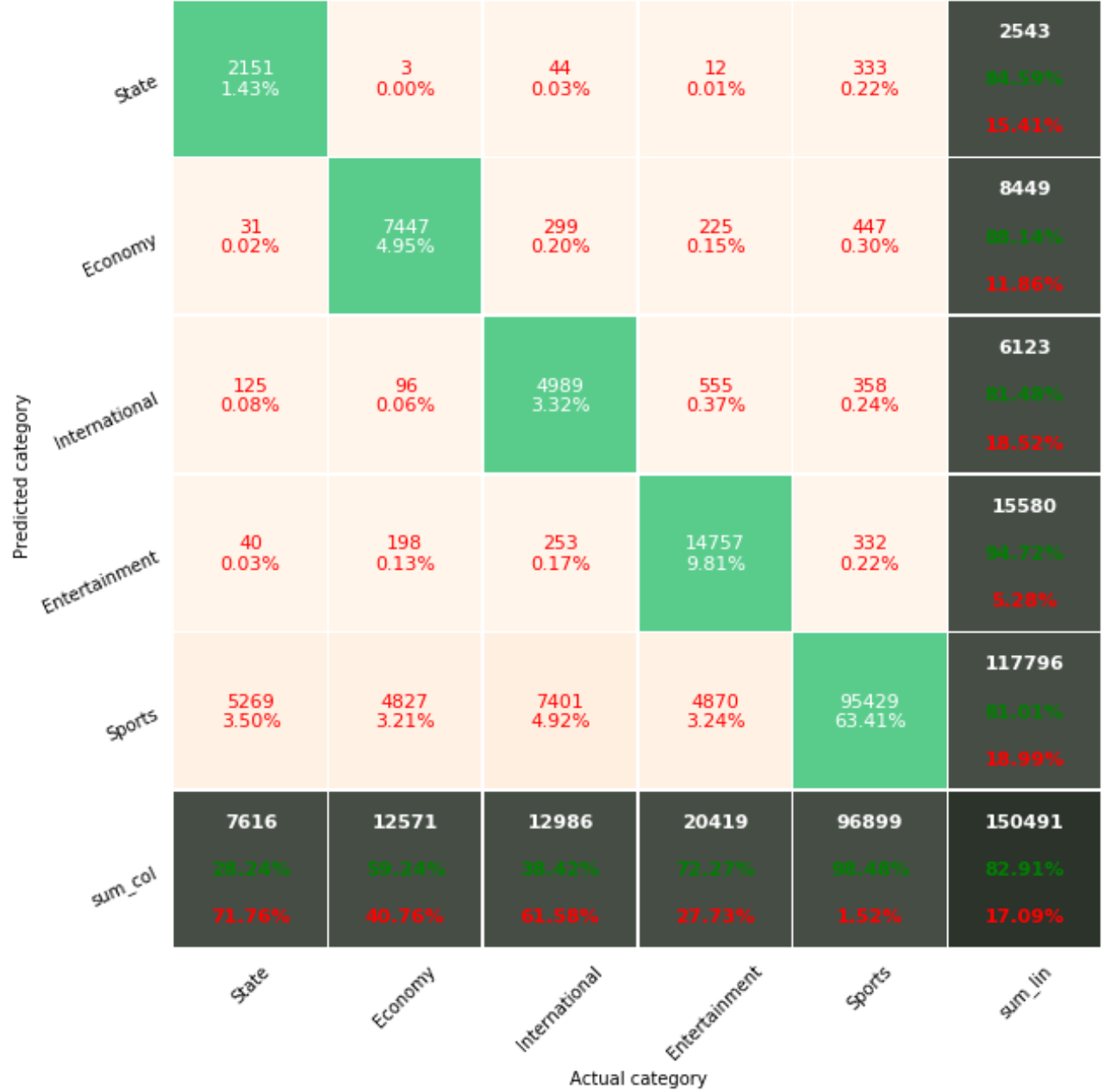}
		\caption{Without pretrained word embeddings}
        \label{fig:confa}
	\end{subfigure}
	\begin{subfigure}{0.48\linewidth}
		\centering
		\includegraphics[width=\linewidth,height=75mm]{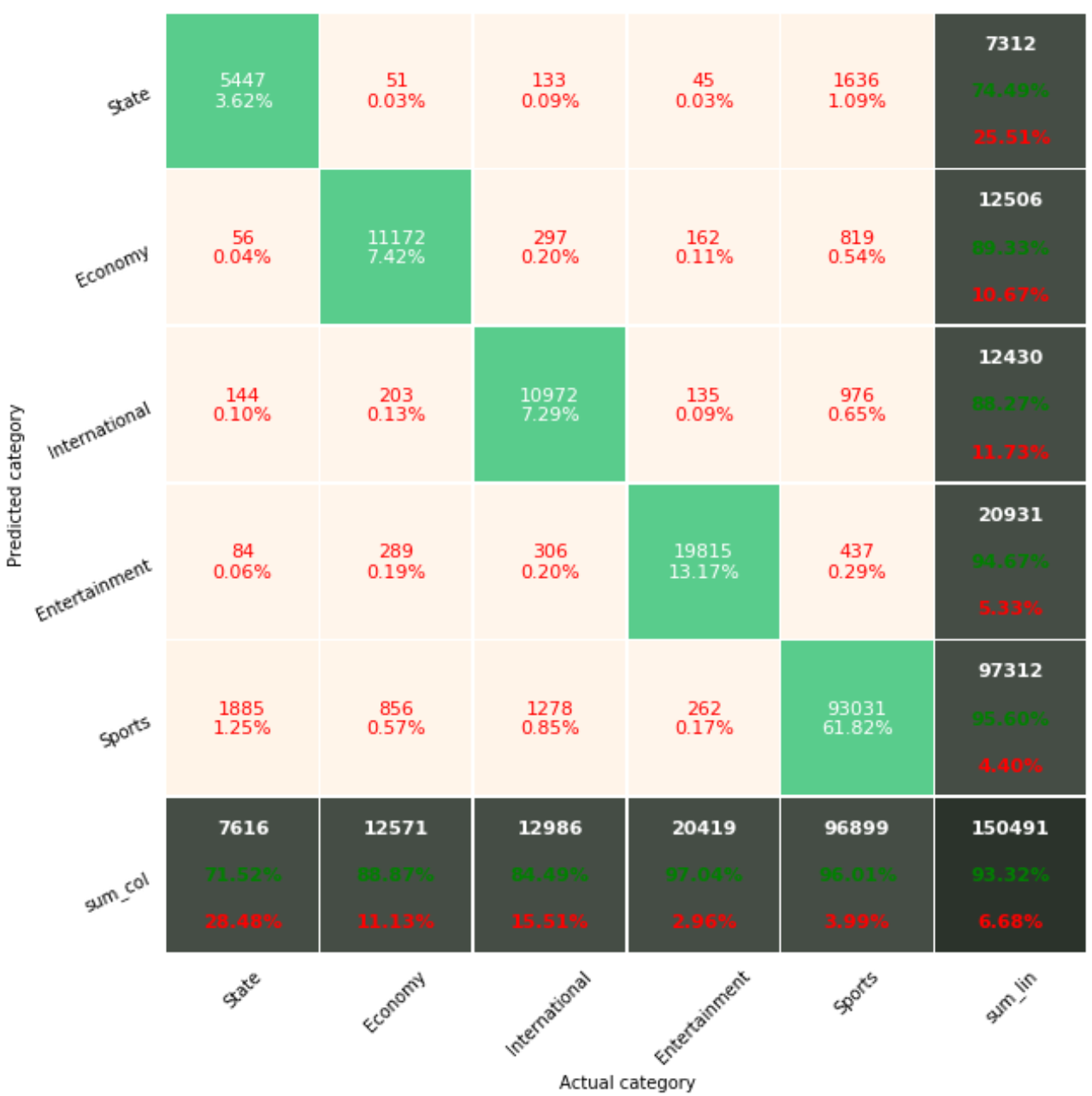}
		\caption{Using pretrained word embeddings}
        \label{fig:confb}
	\end{subfigure}
	\caption{Confusion matrix of the \texttt{MConv-LSTM} model for document classification} 
	\label{fig:confusions}
\end{figure*}

As shown in \cref{fig:confusions}, classical models, specially KNN and NB, often confused economics and international related topics and managed to classify them 45\% and 35\% of cases correctly. 
This is because a document from both these categories contains common words and topics discussing finance, development, share market, intentional affairs, and politics. 
Further, as shown in \cref{table:doc}, the \texttt{MConv-LSTM} model also confused between similar categories plus the sports. However, for state-related topics, both classic and \texttt{MConv-LSTM} classifiers perform quite well.

Each model's performance is then evaluated after initializing network's weights using the pre-trained \texttt{BengFastText} model. Using the latter approach, the accuracy increased by 2.0 to 9.45 points weighted F1~(except NB, which performed even worse with the word embeddings). 
As shown in \cref{table:all}, among standalone classifier, most significant boost was with \texttt{MConv-LSTM} by 8.75\%, which is about 12\% improvement compared to the previous results. Overall, MAE gives 9.45\% boost, which is 1.3\% better than the second best \texttt{MConv-LSTM} showing robustness of our approach.

\subsection{\textbf{Analysis of hate speech detection}} 
\label{subsection5}
We evaluated the model with and without initializing network weight using pretrained \texttt{BengFastText} model. Using the pretrained word embeddings, the overall accuracy is increased by 2.15 to 7.27 points weighted F1 as shown in \cref{table:hate_result}. Nevertheless, each classifier's performance improved with the pretrained word embeddings except for the NB~(performs the worst with only 69.3\% accuracy). Among tree-based classifiers, GBT performs the best giving as much as 86.2\% accuracy, which is the best among classic models as well. Overall, MAE gives 7.27\% boost, which is already 1.10\% better than the second best \texttt{MConv-LSTM}~(that performs best as a standalone classifier giving 7.15\% improvement).

\begin{table*}
    \caption{Performance of hate speech detection  based on embedding methods}
    \label{table:hate_result}
    \vspace{-3mm} 
    \scriptsize{
    \begin{tabular}{c|l|l|l|l}
        \hline
        \textbf{Embedding method} & \textbf{Classifier} & \textbf{Precision} & \textbf{Recall} & \textbf{F1}\\ \hline
         \multirow{7}{*}{\textbf{BengFastText}} & LR & 0.723+2.21\% &	0.727+2.25\% &	0.723+2.15\% \\
         & NB & 0.691-1.05\% &	0.694 -1.12\% &	0.693 - 1.15\% \\
         & SVM & 0.735+2.54\% &	0.729+2.90\% &	0.728+2.25\% \\
         & KNN & 0.716+2.61\% &	0.711+2.52\% &	0.712+2.32\% \\
         & GBT & 0.842+5.37\% &	0.845+5.41\% &	0.845+5.79\%\\
         & RF & 0.861+6.25\% &	0.857+4.91\% &	0.862+5.32\% \\
         & MConv-LSTM  & 0.881+\color{blue}{7.25\%} &	0.883+\color{blue}{7.54\%} &	0.882+\color{blue}{7.15\%}  \\
         & MAE & 0.894+\color{blue}{7.45\%} &	0.896+\color{blue}{7.65\%} &	0.891+\color{blue}{7.27\%} \\ \hline
         \multirow{7}{*}{\textbf{GloVe}} & LR & 0.718+1.54\% &	0.715+1.65\% &	0.716+1.40\% \\
         & NB & 0.672-2.21\% &	0.673 -2.12\% &	0.681 - 2.15\% \\
         & SVM & 0.724+2.92\% &	0.726+2.75\% &	0.726+2.65\% \\
         & KNN & 0.715+2.7\% &	0.716+2.75\% &	0.714+2.53\% \\
         & GBT & 0.823+4.772\% &	0.825+5.11\% &	0.824+5.17\%\\
         & RF & 0.818+5.93\% &	0.824+4.72\% &	0.821+6.10\% \\
         & MConv-LSTM  & 0.827+\color{blue}{6.12\%} &	0.822+\color{blue}{6.12\%} &	0.824+\color{blue}{6.11\%}  \\
         & MAE & 0.831+\color{blue}{6.55\%} &	0.834+\color{blue}{6.65\%} &	0.837+\color{blue}{6.9\%} \\ \hline
        \multirow{7}{*}{\textbf{Word2Vec}} & LR & 0.69+1.75\% &	0.70+2.6\% &	0.71+1.25\% \\
         & NB & 0.65-2.5\% &	0.66 -2.2\% &	0.67 - 1.6\% \\
         & SVM & 0.69+2.1\% &	0.70+1.75\% &	0.77+2.6\% \\
         & KNN & 0.70+2.5\% &	0.71+2.6\% &	0.71+2.5\% \\
         & GBT & 0.806+4.854\% &	0.815+4.9\% &	0.810+5.1\%\\
         & RF & 0.74+5.6\% &	0.75+4.7\% &	0.74+5.3\% \\
         & MConv-LSTM  & 0.77+\color{blue}{5.7\%} &	0.78+\color{blue}{5.5\%} &	0.78+\color{blue}{5.8\%}  \\
         & MAE & 0.79+\color{blue}{6.45\%} &	0.80+\color{blue}{6.25\%} &	0.79+\color{blue}{6.7\%} \\ \hline
    \end{tabular}
    }
\end{table*}

\begin{figure*}
	\centering
	\begin{subfigure}{0.5\textwidth}
		\centering
		\includegraphics[width=\linewidth,height=70mm]{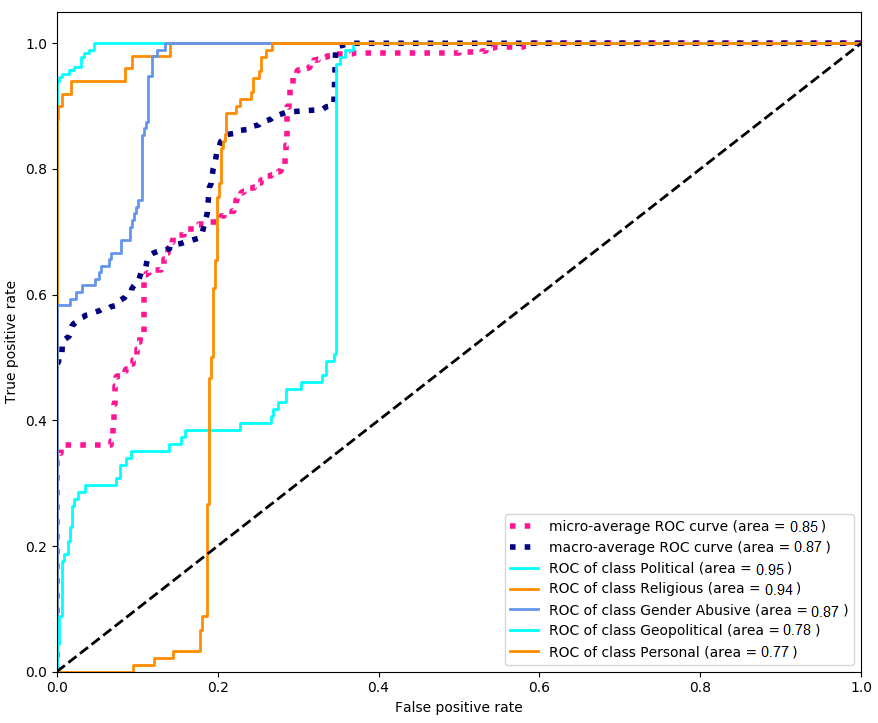}
		\caption{Without pretrained word embeddings}
        \label{fig:roc_hatespeech_without_pretrained}
	\end{subfigure}%
		\begin{subfigure}{0.5\textwidth}
		\centering
		\includegraphics[width=\linewidth,height=70mm]{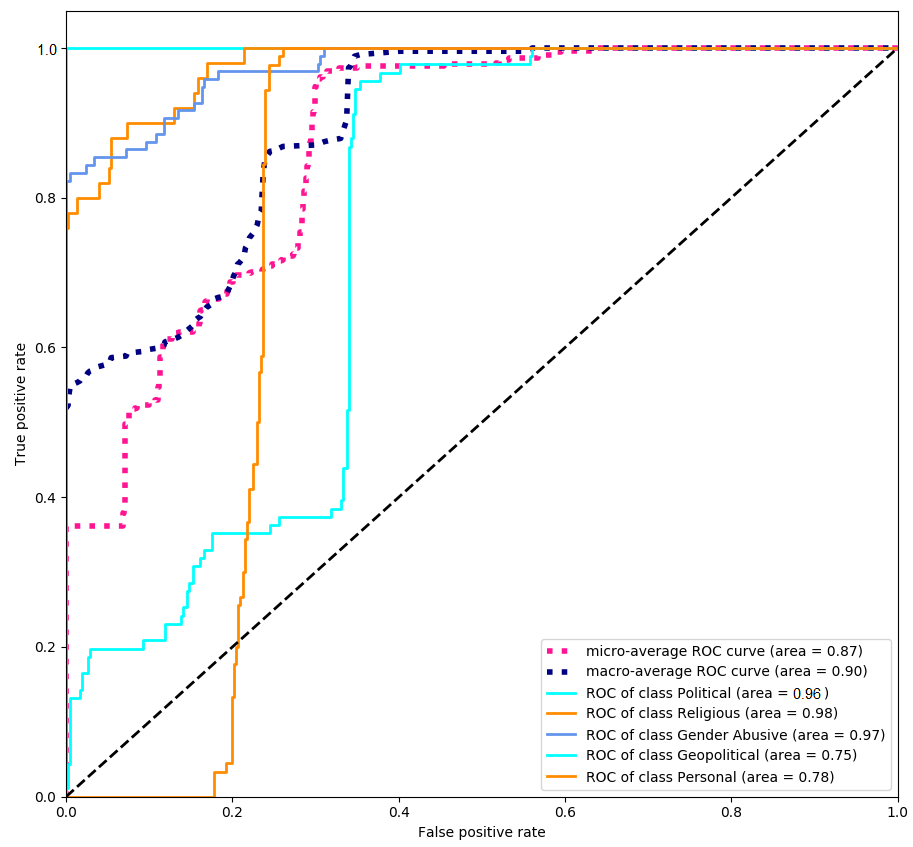}
		\caption{With pretrained word embeddings}
        \label{fig:roc_hatespeech_with_pretrained}
	\end{subfigure}
	\caption{ROC curves of cross-validated \texttt{MConv-LSTM} model(hate speech detection)} 
	\label{fig:roc_curves}
	\vspace{-2mm} 
\end{figure*}

\begin{figure}[h]
	\centering
	\includegraphics[width=0.5\textwidth,height=65mm]{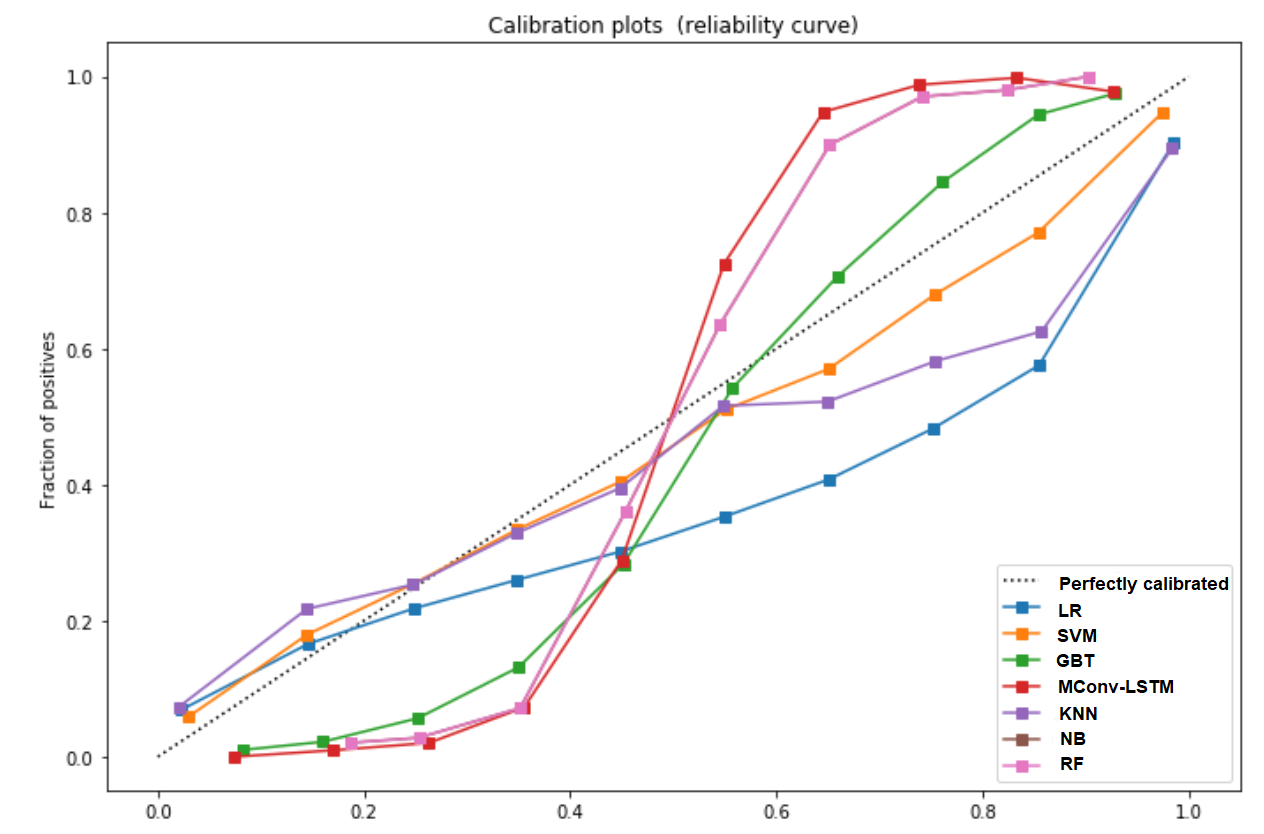}	
    \caption{Calibrating classifiers on word embeddings using \texttt{BengFastText} showing reliability curves}
	\label{fig:cal}
\end{figure}

This analysis can be further validated by calibrating the best performing Conv-LSTM classifier against different embedding methods for which the output probability of the classifier can be directly interpreted as a confidence level in terms of `fraction of positives' as shown in~\cref{fig:cal}. As seen the Conv-LSTM classifier gave a probability value between 0.82 to 0.93, which means 93\% predictions belong to true positive predictions generated by the \texttt{BengFastText}. 

\subsection{\textbf{Performance of sentiment analysis}} 
\label{sec:performance-sentiment-analysis}
\Cref{table:sentiment_result} summarizes the results of sentiment analysis, which again suggest that each model performs better with the pre-trained word embeddings, measured by the \texttt{MCC} showing strong positive relationships. This result suggests that the predictions were strongly correlated with the ground truth, yielding a Pearson product-moment correlation coefficient higher than 0.65 for all the classifiers. This is because \texttt{MConv-LSTM} model learns better with the pre-trained word embeddings, which why it gives about 5.49\% boost compared to without pre-trained word embeddings in terms of AUC score. Overall, the MEA gives 7.85\% improvement. 

\begin{table}
    \caption{Performance of sentiment analysis}
    \label{table:sentiment_result}
    \vspace{-3mm} 
    \scriptsize{
    \begin{tabular}{c|l|l|l}
        \hline
        \textbf{Embedding method} & \textbf{Classifier} & \textbf{MCC} & \textbf{AUC} \\ \hline
         \multirow{7}{*}{\textbf{BengFastText}} & LR & 0.652+2.9\% &	0.76+2.8\% \\
         & NB & 0.653 + 3.25\% &	0.77 + 2.32\% \\
         & SVM & 0.694+2.25\% &	0.79+2.45\% \\
         & KNN & 0.701+2.12\% &	0.80+3.33\% \\
         & GBT & 0.715+3.18\% &	0.85+3.91\% \\
         & RF & 0.727+3.67\% &	0.86+4.26\%  \\
         & MConv-LSTM  & 0.746+\color{blue}{5.49\%} &	0.87+\color{blue}{7.13\%}  \\
         & MAE & 0.758+\color{blue}{7.85\%} &	0.88+\color{blue}{7.42\%} \\ \hline
         \multirow{7}{*}{\textbf{GloVe}} & LR & 0.646+2.9\% &	0.75+1.75\% \\
         & NB & 0.644 + 2.10\% &	0.76 + 2.11\% \\
         & SVM & 0.681+1.85\% &	0.78+2.51\% \\
         & KNN & 0.695+2.22\% &	0.79+2.28\% \\
         & GBT & 0.707+2.63\% &	0.84+2.67\% \\
         & RF & 0.719+2.45\% &	0.85+3.17\%  \\
         & MConv-LSTM & 0.736+\color{blue}{4.76\%} &	0.86+\color{blue}{5.78\%}  \\
         & MAE & 0.734+\color{blue}{6.62\%} &	0.87+\color{blue}{5.29\%} \\ \hline
         \multirow{7}{*}{\textbf{Word2Vec}} & LR & 0.634+2.6\% &	0.73+2.7\% \\
         & NB & 0.645 + 2.25\% &	0.71 + 2.40\% \\
         & SVM & 0.677+1.75\% &	0.77+1.5\% \\
         & KNN & 0.662+3.2\% &	0.78+3.3\% \\
         & GBT & 0.683+3.5\% &	0.83+3.8\% \\
         & RF & 0.692+4.5\% &	0.82+4.3\%  \\
         & MConv-LSTM  & 0.705+\color{blue}{6.4\%} &	0.83+\color{blue}{6.5\%}  \\
         & MAE & 0.716+\color{blue}{8.1\%} &	0.85+\color{blue}{7.4\%} \\ \hline
    \end{tabular}
    }
\end{table}

On the other hand, AUC score of the classifier found to be the highest is also generated by the \texttt{MConv-LSTM} network, which is at least 2\% better than the second-best score by the RF classifier but the worst performance was recorded by the LR classifier. The ROC curves generated by the \texttt{MConv-LSTM} model are shown in \cref{fig:roc_hatespeech_without_pretrained} and \cref{fig:roc_hatespeech_with_pretrained}. As seen AUC scores are consistent across folds and This signifies that the predictions by the \texttt{MConv-LSTM} model are much better than random guessing. 


\subsection{\textbf{Effects of feature selection}} 
Based on the results for each classification task, it is obvious that feature selection can be a very powerful technique to improve the performance of classic methods especially with NB, SVM, KNN, and tree-based approaches. 
Contrarily, although LR is intrinsically simple, has low variance, and less prone to over-fitting, feature selector based on this may be discriminating features very aggressively. This forces the model to lose some useful features, which results in lower performance. While investigating alternative feature selection algorithms, e.g., recursive feature elimination or genetic algorithms is beyond the scope of this research. 

\begin{figure*}[htp!]
	\centering
	\begin{subfigure}{0.48\linewidth}
		\centering
		\includegraphics[width=\linewidth,height=55mm]{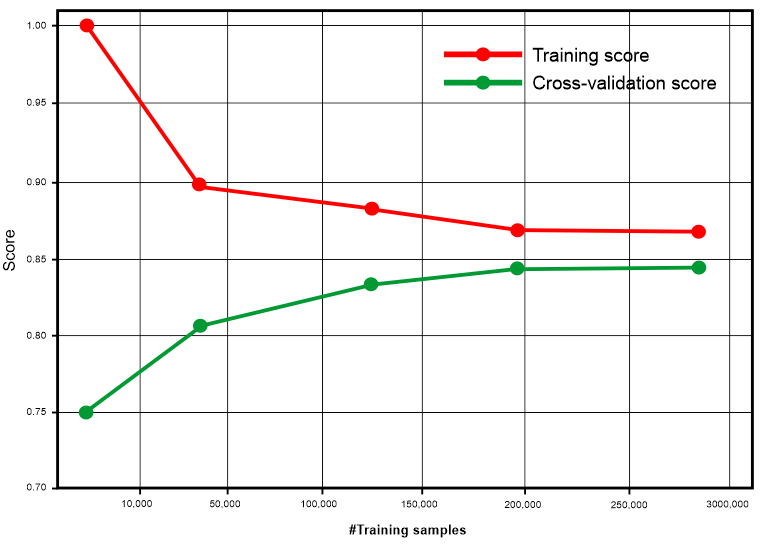}
		\caption{SVM}
        \label{fig9a}
	\end{subfigure}
	\begin{subfigure}{0.48\linewidth}
		\centering
		\includegraphics[width=\linewidth,height=55mm]{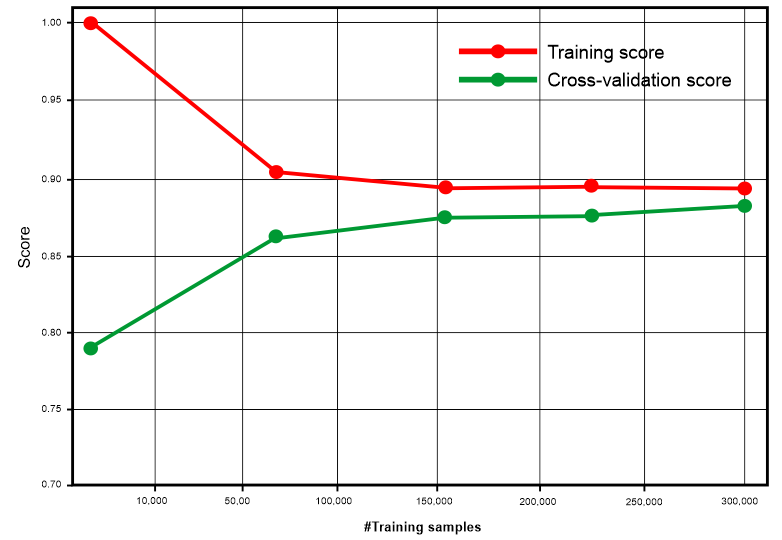}
		\caption{GBT}
        \label{fig9b}
	\end{subfigure}
	\begin{subfigure}{0.48\linewidth}
		\centering
		\includegraphics[width=\linewidth,height=55mm]{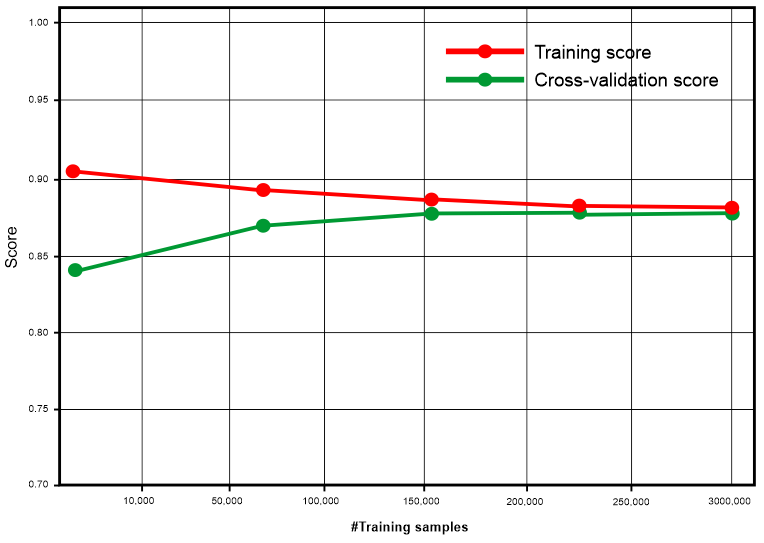}
		\caption{RF}
        \label{fig9c}
	\end{subfigure}
		\begin{subfigure}{0.48\linewidth}
		\centering
		\includegraphics[width=\linewidth,height=55mm]{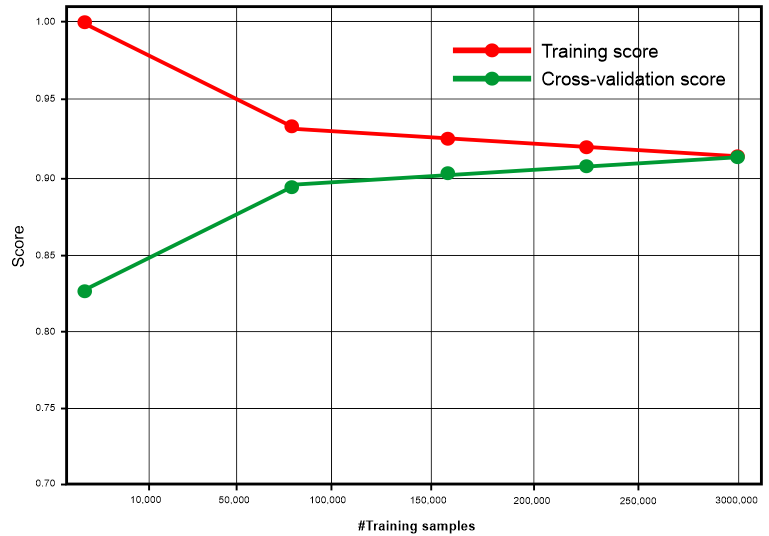}
		\caption{Conv-LSTM}
        \label{fig9d}
	\end{subfigure}
	\caption{Learning curves with validation and training scores of top classifiers} 
	\label{fig:lc}
\end{figure*}

Another reason could be the size of the training data, which is comparatively small, i.e., hate speech dataset. However, tree-based approaches such as RF and GBT perform quite better compared to NB, SVM, and KNN, especially, NB which even performs worse with pre-trained word embeddings. One potential reason behind such worse performance of NB is that it's conditional independence assumption,which assumes features are independent of one another when conditioned upon class labels, was rarely accurate in our case. Compared to classic models, neural network based \texttt{MConv-LSTM} models seems to be much better (about 3 to 4\%) as it outperforms the best performing classic GBT model in F1-score, which is probably because neural networks managed to capture abstract features that couldn't be modeled by classic models. 

\subsection{Effects of number of samples}
To understand the effects of having more training samples, and to understand if our classifiers suffer more from variance errors or bias errors, we observed the learning curves of top-3 classifiers~(i.e., RF, GBT, and \texttt{MConv-LSTM}) and SVM~(a linear model) for varying numbers of training samples in case of document classification. 

As shown in \cref{fig:lc}, both the validation and training scores converge to a value, which is too low with increasing size of the training set. Hence, SVM didn't benefit much from more training samples. 
However, since RF and GBT are tree-based ensemble methods and \texttt{MConv-LSTM} model learn more complex concepts producing lower bias, the training scores are much higher than the validation scores for the maximum number of samples, i.e., adding more training samples has increased model's generalization. 
\subsection{Comparison with baseline embeddings}
Our empirical study founds that almost every classifier performs poorly across every supervised learning task that are based on the embeddings generated by \texttt{BengFastText} method compared to Word2Vec and GloVe, as shown in \cref{table:all}, \cref{table:sentiment}, and \cref{table:hate}. In particular, except for the sentiment analysis task in which classifiers performance were consistent, whereas each classifier's performance were clearly boosted based on embeddings generated by the \texttt{BengFastText} model, whereas. Further, an initial experiment based on Polyglot embedding method shows that it failed to capture the semantic contexts from texts, which is probably because Polyglot uses only 64 dimensions for the embeddings. 

Although classifiers trained on embeddings from Word2Vec model work well across tasks, the online scanning approach used by Word2Vec is sub-optimal since the global statistical information regarding word co-occurrences does not fully exploited~\cite{GloVe}. GloVe, on the other hand, performed better than that of Word2Vec since it produces a vector space with meaningful substructure and performs consistently well in word analogy tasks and outperforms related models on similarity tasks and named entity recognition~\cite{GloVe}.

Further, Word2Vec and GloVe both failed to provide any vector representation for words that are not in the model dictionary, i.e., not resilient against OOV. Hence, the approach used to tackle the OOV in our approach~(i.e., randomly assigned word out of in-vocabulary words list from the pretrained model by considering the surrounding-contexts) won't give accurate embeddings~\cite{bojanowski2017enriching}. On the other hand, fastText performed much better compared to other embedding methods since fastText works well with rare words. Thus, even if a word was not seen during the training, it can be broken down into n-grams to get it's corresponding embeddings. 

\section{Conclusion and outlook}
\label{con}
In this paper, we provide three different classification benchmarks for under-resourced Bengali language, which is based on word embeddings and a deep neural network architecture called \texttt{MConv-LSTM}. The largest \texttt{BengFastText} word embedding model, which is built on 250 million Bengali articles seems able to sufficiently capture the semantics of words. 
To show the effectiveness of our approach, we prepared three datasets of hate speech, sentiment analysis, and commonly discussed topics. Then we experimented on three use cases: hate speech detection, sentiment analysis, and text classification, using a method that combines CNN and LSTM networks with dropout and pooling, which is found to empirically improve classification accuracy across all use cases. 
We also conducted comparative evaluations on all of these datasets and show that the proposed method outperformed classic models across metrics and use cases. 

Nevertheless, we provided the results based on other Bengali language embedding methods like fastText and Polyglot. Our results show that approaches based on \texttt{BengFastText} embeddings clearly outperformed Word2Vec and GloVe-based ones. 
Classic methods that depend on pre-engineered features do not require sophisticated feature engineering but using automatic feature selection techniques on generic features e.g. n-grams can, in fact, produce better results.
As research suggests character-based features to be more effective than word-based ones, taking this into account, we experienced better results with \texttt{MConv-LSTM} using only word-based features, which is likely due to the superiority in the network architecture. Results of different experiments showed that feature selection with drop-out, Gaussian noise layer, and pooling can significantly enhance the learning capabilities to \texttt{MConv-LSTM} towards through better training and learning. 

Research also found that misspellings of words often appeared in up to 15\% of web search queries~\cite{piktus2019misspelling}. Such misspellings words are also responsible for the OOV~(apart from other issues like abbreviations, slang, etc.) in language models. Likewise, a recent language model called Misspelling Oblivious Word Embeddings~(MOWE)~\cite{piktus2019misspelling} in the context of Bengali language to generate word embeddings. MOWE, which is a combination of fastText and a supervised learning technique tries to embed misspelled words close to their correct context and variants. This makes MOWE resilient to misspellings.

In the future, we intend to employ MOWE-based approach to generate the embeddings and explore other possibilities such as i) building and training a likely more robust neural network by stacking multiple conv and bidirectional-LSTM layers, which are good for extracting hierarchical features, ii) using Bidirectional Encoder Representations from Transformers~(BERT)~\cite{devlin2018BERT}~(based on deep bidirectional representations from unlabeled text by jointly conditioning on both left and right context in all layers) for solving similar problem even without using word embeddings, iii) integrating user-centric features, e.g., frequency of a user detected for posting hate speech and the user’s interaction with others, iv) studying and quantifying the difference between hate speech, abusive or offensive language, and cyber-bullying, v) further break sentiment polarities down into specific categories, e.g., sentiment toward Cricket, electronics, food, or celebrities.   

\section*{Acronyms}
\noindent Acronyms and their full forms used in this paper are as follows:
\vspace{-3mm}

\begin{multicols}{2}
\begin{description}[leftmargin=0pt]
\scriptsize{
    \item [BERT] Bidirectional Encoder Representations from Transformers
    \item [CNN] Convolutional Neural Networks
    \item [CSPlang] Creative Political Slang
    \item [CRF] Conditional Random Field
    \item [DT] Decision Trees
    \item [DL] Deep Learning 
    \item [GBT] Gradient Boosted Trees
    \item [GRU] Gated Recurrent Unit
    \item [LR] Logistic Regression 
    \item [LSTM] Long Short-term Memory
    \item [MConv-LSTM] Multichannel Convolutional-LSTM
    \item [ML] Machine Learning
    \item [MCC] Matthias Correlation Coefficient
    \item [MAE] Model Averaging Ensemble
    \item [MOWE] Misspelling Oblivious Word Embeddings
    \item [NER] Named Entity Recognition
    \item [NB] Na{\"i}ve Bayes
    \item [NLP] Natural Language Processing
    \item [OOV] Out-of-vocabulary
    \item [PoS] Part-of-speech
    \item [RF] Random Forest 
    \item [ReLU] Rectified Linear Unit
    \item [SVM] Support Vector Machines.
	 }
\end{description}
\end{multicols}

\bibliographystyle{ACM-Reference-Format}
\bibliography{Main}

\end{document}